\documentclass[10pt]{article} % For LaTeX2e
\usepackage[accepted]{rlc}
\usepackage{amssymb}            % Defines common symbols like \mathbb R
\usepackage{mathtools}          % Extends amsmath, providing common math tools
\usepackage{mathrsfs}           % Enables \mathscr, which can work in cases that \mathcal does not
\mathtoolsset{showonlyrefs}     % Only number equations that are referenced (optional)
\usepackage{graphicx}           % For including images
\usepackage{subcaption}         % Allows for the use of subfigures and subcaptions
\usepackage[space]{grffile}     % For spaces in image names
\usepackage{url}                % For displaying urls
\usepackage{booktabs}
\usepackage{wrapfig}

\usepackage[toc]{glossaries}
\glsdisablehyper  % Disable hyperlinks for glossary entries
\makeglossaries
\newacronym{rl}{RL}{Reinforcement Learning}

\newacronym{marl}{MARL}{Multi-Agent Reinforcement Learning}

\newacronym{mdp}{MDP}{Markov Decision Processes}
\newacronym{pomdp}{POMDP}{Partially Observable Markov Decision Processes}
\newacronym{dec-pomdp}{Dec-POMDP}{Decentralised Partially Observable Markov Decision Processes}
\newacronym{dec-mdp}{Dec-MDP}{Decentralised Markov Decision Processes}
\newacronym{pomg}{POMG}{Partially Observable Markov Game}
\newacronym{ctde}{CTDE}{Centralised-Training Decentralised-Execution}

\newacronym{gnn}{GNN}{Graph Neural Network}

\newacronym{cnn}{CNN}{Convolutional Neural Network}

\newacronym{sota}{SOTA}{state-of-the-art}

\newacronym{mpc}{MPC}{Model Predictive Control}

\newacronym{mbpo}{MBPO}{Model-Based Policy Optimisation}

\newacronym{mambpo}{MAMBPO}{Multi-Agent Model-Based Policy Optimisation}

\newacronym{aorpo}{AORPO}{Adaptive Opponent-wise Rollout Policy Optimisation}

\newacronym{ppo}{PPO}{Proximal Policy Optimisation}

\newacronym{mamba}{MAMBA}{Multi-Agent Model-Based Approach}

\newacronym{maci}{MACI}{Multi-Agent Communication through Imagination}

\newacronym{intshare}{IS}{Intention Sharing}

\newacronym{mbvd}{MBVD}{Model-Based Value Decomposition}

\newacronym{rssm}{RSSM}{Recurrent State-Space Model}

\newacronym{vmas}{VMAS}{Vectorised Multi-Agent Simulator}

\newacronym{iqm}{IQM}{Interquartile Mean}

\newacronym{trpo}{TRPO}{Trust Region Policy Optimisation}

\newacronym{gat}{GAT}{Graph Attention Network}

\newacronym{ci}{CI}{Confidence Interval}

\newacronym{idreamer}{IDreamer}{Independent Dreamer}

\newacronym{codreamer}{CoDreamer}{Communicative Dreamer}
\title{CoDreamer: Communication-Based Decentralised World Models}

\author{Edan Toledo  \\
    e.toledo@instadeep.com \\
    InstaDeep\\
    \And
    Amanda Prorok \\
    asp45@cam.ac.uk\\
    Department of Computer Science and Technology \\
    University of Cambridge \\}

\begin{document}

\maketitle

\begin{abstract}
Sample efficiency is a critical challenge in \acrlong{rl}. Model-based \acrshort{rl} has emerged as a solution, but its application has largely been confined to single-agent scenarios. In this work, we introduce \acrshort{codreamer}, an extension of the Dreamer algorithm for multi-agent environments. \acrshort{codreamer} leverages Graph Neural Networks for a two-level communication system to tackle challenges such as partial observability and inter-agent cooperation. Communication is separately utilised within the learned world models and within the learned policies of each agent to enhance modelling and task-solving. We show that \acrshort{codreamer} offers greater expressive power than a naive application of Dreamer, and we demonstrate its superiority over baseline methods across various multi-agent environments.
\end{abstract}

%%%%%%%%%%%%%%%%%%%%%%%%%%%%%%%%%%%%%%%%%%%%%%%%%%%%%%%%%%%%%%%%
%% Section: Submission of papers to RLC
%%%%%%%%%%%%%%%%%%%%%%%%%%%%%%%%%%%%%%%%%%%%%%%%%%%%%%%%%%%%%%%%
\section{Introduction}
Reinforcement Learning (RL) has become a leading paradigm for creating autonomous control systems. Despite recent successes \citep{Degrave2022, luo2022controlling, roy2021prefixrl, mirhoseini2021graph, Schrittwieser2020}, these achievements often require vast data and computational resources. While computational power is likely to increase, data availability remains a constraint. As we tackle more complex problems requiring expensive simulations, improving the sample efficiency of RL methods is critical.

Model-based RL \citep{sutton2018reinforcement} offers a promising solution to this issue. Recent algorithms like EfficientZero \citep{ye2021mastering} and Dreamer \citep{Hafner2020Dream, hafner2021mastering, hafner2023mastering} show that high performance can be achieved utilising far less data. These methods develop a world model, a learned simulation of the agent's environment, to generate synthetic data and/or plan actions based on future predictions. However, their effectiveness in multi-agent settings is limited by the accuracy of these models and their single-agent design.

Transitioning to multi-agent systems introduces new challenges such as partial observability and non-stationarity, complicating the use of many single-agent RL algorithms \citep{nguyen2020deep}. Sample efficiency issues become more pronounced when multiple agents need to learn a shared set of policies. This work aims to adapt the success of single-agent model-based RL to multi-agent environments.

We propose \acrfull{codreamer}, an enhancement to the Dreamer algorithm, addressing challenges like partial observability and non-stationarity in multi-agent environments. CoDreamer employs a two-level communication system: agents communicate within their world models to better model their environment, and within their policies to enhance cooperation and performance. This dual-tiered strategy aims to overcome the limitations of existing methods and advance model-based RL in multi-agent scenarios.

As agents use their world models during execution, decentralised communication is essential. Recent research \citep{li2020graph, tolstaya2020learning, prorok2018graph, kortvelesy2022qgnn} shows that Graph Neural Networks (GNNs) can enhance communication and performance in multi-agent systems within the CTDE framework. CoDreamer utilises GNNs for a learned communication strategy, allowing agents to collaboratively generate synthetic trajectories, thereby improving performance given a limited sample budget.

This work makes several key contributions: the implementation and evaluation of the DreamerV3 \citep{hafner2023mastering} algorithm in a multi-agent independent learning setting, termed IDreamer; the development of CoDreamer, an enhanced version of IDreamer that uses GNNs for decentralised communication among agents' world models and policies, addressing issues like non-stationarity and partial observability; and a comprehensive evaluation of CoDreamer across various environments, demonstrating superior performance and more accurate world models in environments with inter-agent dependencies.

\section{Methodology}
\subsection{Independent Dreamer}
To thoroughly examine the effects of communication within world models, we first assess the performance of fully decentralised non-communicative world models. This leads to the creation of distinct Dreamer agents, each using its own separate world model. We introduce IDreamer, which implements DreamerV3 \citep{hafner2023mastering} for the multi-agent setting.

Like most independent multi-agent RL (MARL) algorithms, each IDreamer agent treats other agents as part of the environment, with no explicit communication or opponent modelling. Each agent trains solely on its own experiences and does not observe other agents during training or execution.

During actor-critic network training, agents use only their own observations to start trajectory imagination. Due to the substantial size of the world model networks (16M+ parameters), we use parameter sharing \citep{gupta2017cooperative} to expedite learning. Although parameter sharing allows the world model to train on all agents' experiences, it is still conditioned on individual agents' observations, without additional information for trajectory imagination or observation encoding.

Given the homogeneous nature of the agents and parameter sharing, the world model may struggle to distinguish which agent it is being used by. To address this, we concatenate agent indices as a one-hot encoding to all observations in non-visual environments, enabling the world model to differentiate between agents. We hypothesise that in visual environments, there is sufficient information for the model to distinguish agents.

Apart from these multi-agent specific changes, the implementation details of IDreamer remain unchanged from DreamerV3.

\subsection{CoDreamer: Communicative World Models}
Using world models independently can exacerbate the challenges of multi-agent systems. In direct-style model-based methods like IDreamer, each policy is trained solely with data from its agent's world model, making policy performance entirely dependent on the model's accuracy. Issues like non-stationarity, partial observability, and cooperation are worsened if the model fails to capture other agents' actions. Independent world models struggle as the environment dynamics seem to constantly change, and single-agent observations and actions are insufficient to infer future information, potentially hindering accurate model development.

Moreover, if the environment's reward and transition functions are highly inter-agent dependent, learning through imagination becomes impossible because the imagined trajectories will be inaccurate. To address these issues while staying within the CTDE framework, we propose CoDreamer.

CoDreamer enhances IDreamer by incorporating two levels of communication. The first level, used and learned by the world models, improves trajectory imagination among agents, helping with non-stationarity and partial observability. This communication is independent of actor-critic learning and focuses solely on better state representations and environment modelling. The world models learn communication grounded in the specific environment rather than any agent's policy.

The second level of communication is dedicated to the actor and critic networks trained during imagination, enabling agents to share action and value prediction information. This level is consistent with communication in other MARL methods, focusing on action-relevant information.

For both communication levels, we use GNNs due to their applicability in both centralised and decentralised contexts. GNNs, in our case the GAT V2 architecture \citep{brody2022how}, enable $k$-hop aggregation of nodes, allowing all agents to share information as long as each can communicate with at least one other agent.

\subsubsection{Communication}
As is consistent with the literature \citep{kortvelesy2022qgnn, li2020graph, blumenkamp2021emergence}, to learn communication between agents using \acrshort{gnn}s, we model the inter-agent communication with a graph $\mathcal{G} = \langle \mathcal{V}, \mathcal{E} \rangle$, where each node $i \in \mathcal{V}$ represents an individual agent $i \in n$, and each edge $e^{ij} \in \mathcal{E}$ represents a communication link between agents $i$ and $j$. The adjacency matrix $\mathbf{A}$ establishes the edge set for each agent, which is contingent on the agent's communication range $C$ set by the environment. In this work, adjacency matrices are constructed using Euclidean distance. Specifically, given two agents $i$ and $j$ with positions $p^i$ and $p^j$, an edge is created if $ ||p^i-p^j|| \leq C $. This communication range is dependent on the environment and other limiting factors. As the current state $s \in \mathcal{S}$ changes over time, $\mathbf{A}$ adapts dynamically based on the changing positions of the agents. The set of neighbouring agents $\mathcal{N}$ that are capable of communicating with agent $i$ is defined as $\mathcal{N}^i = \{v^j | e^{ij} \in \mathcal{E}\}$. Additionally, the Euclidean distance of each edge $e^{ij}$ is stored in a matrix $\mathbf{E} \in \mathbb{R}^{|\mathcal{V}| \times |\mathcal{V}|}$ that can be utilised by the \acrshort{gnn}s.

\subsubsection{World Models}
\begin{figure}[h!]
    \centering
    \includegraphics[width=\linewidth]{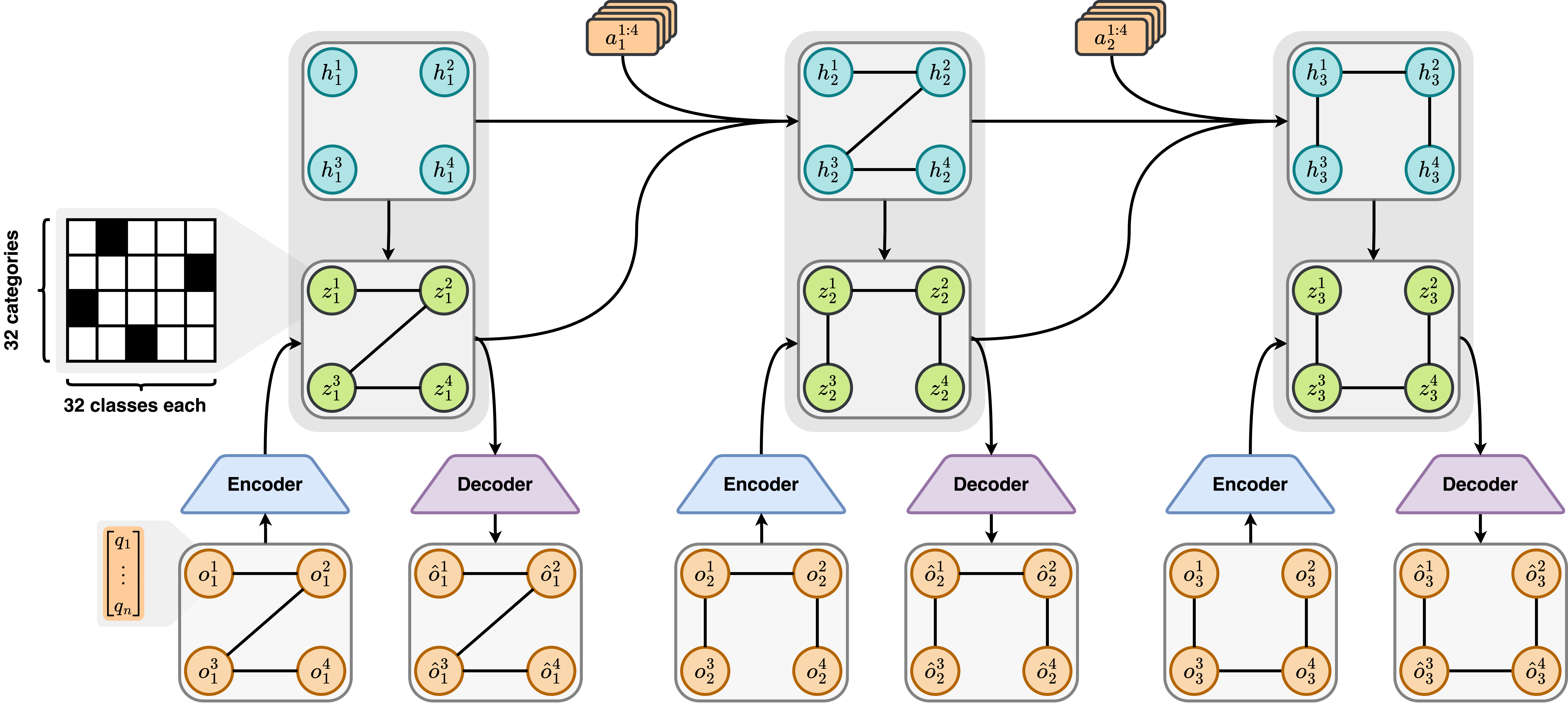}
    \caption[Training process of \acrshort{codreamer} world model]{\textbf{Training process of \acrshort{codreamer} world model:} \acrshort{codreamer} trains in a similar manner to \acrshort{idreamer} (see Figure \ref{fig:idreamer_wm}). However, \acrshort{codreamer} operates over graphs of observations $G^o_t$ and produces graphs of stochastic posterior $G^z_t$ and prior $\hat{G}^z_t$ states. Additionally, recurrent states are also represented as graphs $G^h_t$. This allows communication to happen at any level within the world model architecture.}
    \label{fig:codream_wm}
\end{figure}
In CoDreamer, a unified world model can be used independently by clusters of agents during execution. All elements from IDreamer remain unchanged except for the addition of $k$ GNN layers within the \acrfull{rssm} and prediction heads to facilitate communication. Specifically, only the reward and terminal state prediction heads use communication, while the decoder head remains independent. Each GNN layer performs node updates and neighborhood aggregation, concatenating the original node features to the output to help the world model distinguish agents and prevent over-smoothing \citep{Li_Han_Wu_2018}.

Although we focus on environments formalised as Dec-POMDPs, CoDreamer is adaptable to various multi-agent environments and formalisms, such as Markov Games \citep{shapley1953stochastic}, due to each agent's ability to independently use their prediction heads for local environment estimates.

Unlike IDreamer, CoDreamer's training involves graphs representing all agents' experiences rather than individual experiences. Depending on the adjacency matrix $\mathbf{A}$, agents can be processed independently, effectively transforming CoDreamer into IDreamer for those agents. All graphs have edge features representing the relative distances between agents, which are used with observations in the GAT V2 to calculate attention coefficients for neighborhood aggregation. Beyond these and architectural modifications, CoDreamer remains consistent with IDreamer.

Figure \ref{fig:codream_wm} shows CoDreamer unrolling its world model on a sequence of graphs where each node represents an agent's observation. 

\subsubsection{Behaviour Learning}
\begin{figure}[h!]
    \centering
    \includegraphics[width=0.7\linewidth]{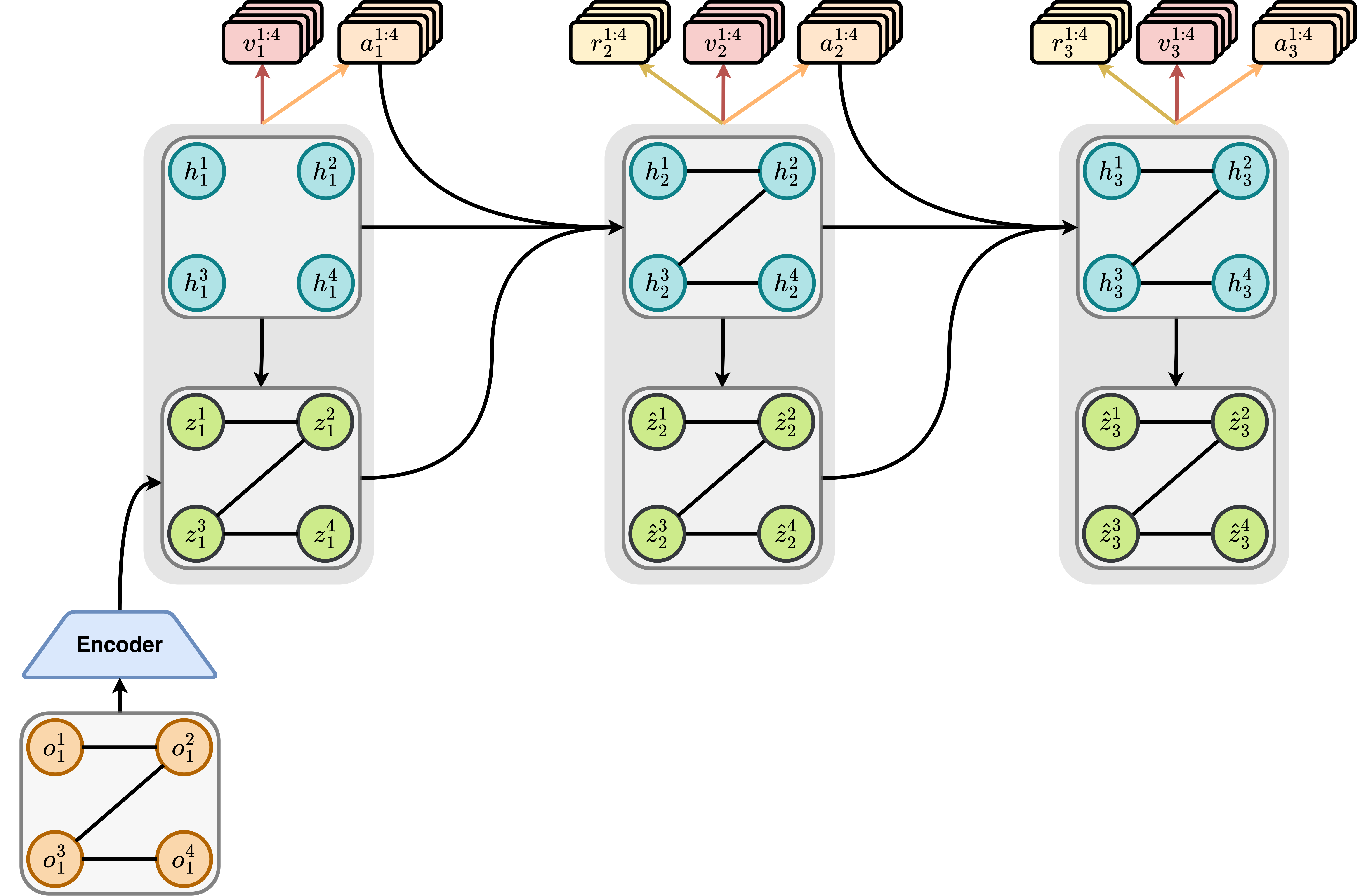}
    \caption[Training process of \acrshort{codreamer} Actor-Critic]{\textbf{Training process of \acrshort{codreamer} Actor-Critic:} To train the actor-critic networks in \acrshort{codreamer}, trajectories of compact world model state graphs $G^{z||h}_{t:H}$ are produced. As \acrshort{codreamer} does not predict new adjacency matrices over the \textit{imagined} trajectories, the adjacency matrix of the starting graph is used for all subsequent graphs. We posit that, as the most relevant agents for short-term future predictions are those nearby at the start, this is sufficient to create more consistent trajectories.}
    \label{fig:codream_policy}
\end{figure}
CoDreamer modifies the behavior learning process similarly to its training procedure. The world model generates trajectories of graphs, with each agent's compact world model state as node features. These imagined trajectories use the adjacency matrix and edge features of the original graph that started the trajectory imagination. We believe the starting graph's adjacency matrix suffices for necessary communication over the imagination horizon, as relevant agent information for future prediction likely depends on agents within the initial communication range. However, original edge features may become outdated over time. Updating these features every step with the GNN could be a future improvement. Each agent uses the predicted graphs to train their actor and critic networks. Figure \ref{fig:codream_policy} illustrates CoDreamer's modifications to IDreamer's behavior learning procedure.

For CoDreamer's synthetic data generation at training time, following the CTDE framework, we could use fully connected adjacency matrices during the imagination rollout. However, since the second level of communication is trained on the imagined graphs, realistic communication topologies must be maintained. Otherwise, at execution time, the actor-critic networks won't effectively communicate relevant information with a non-fully connected adjacency matrix.

\begin{wrapfigure}{r}{0.5\textwidth}
    \centering
    \includegraphics[width=0.6\linewidth]{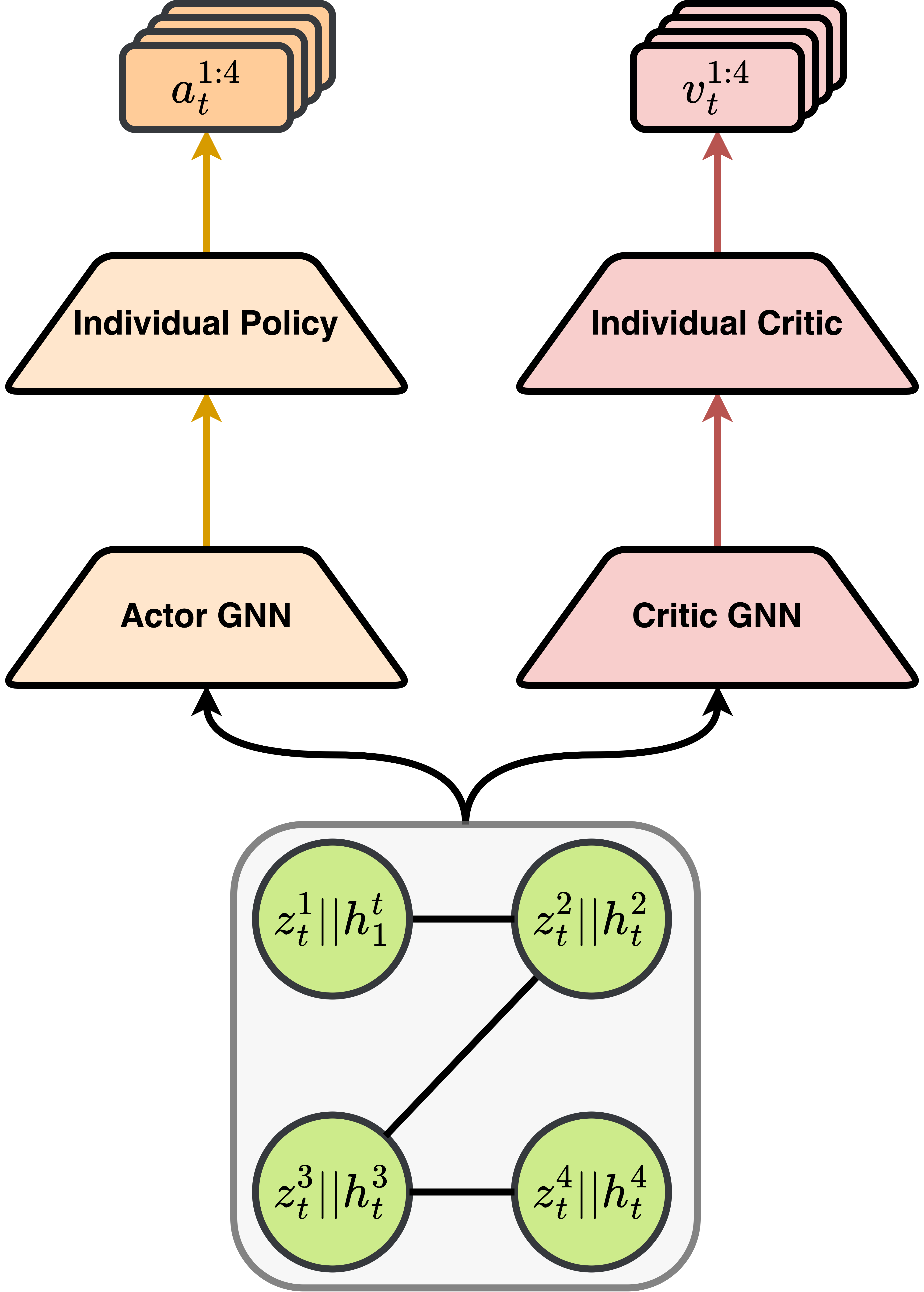}
    \caption[Illustration of \acrshort{codreamer} Actor-Critic]{\textbf{Illustration of \acrshort{codreamer} Actor-Critic:} The \acrshort{codreamer} actor-critic networks operate over the \textit{imagined} graphs of world model states. This second level of communication is trained to maximise the reward and thus exchange relevant information that is not specific to environment modelling.}
    \label{fig:commpolicy}
\end{wrapfigure}

In addition to the state information communicated by the world model, agents also communicate relevant information to learn their policy and value functions. Both actor and critic networks, operating on compact world model state graphs, have distinct GNN layers acting as a torso for the respective networks. Figure \ref{fig:commpolicy} illustrates how the compact world model graphs are used by the policy and value functions during training and execution. Beyond the architectural changes, the actor-critic networks are trained in the same way as IDreamer.

Through the utilisation of an intra-model communication layer, \acrshort{codreamer} is afforded a higher level of expressivity in modelling various environments, compared to \acrshort{idreamer}. Unlike \acrshort{idreamer}, which relies on an independent modelling approach, \acrshort{codreamer} overcomes the significant limitation of not being able to capture inter-agent dependent transition dynamics and reward functions. \acrshort{codreamer} is strictly more expressive in that it can model the true underlying reward and transition functions of \acrshort{dec-mdp}s \citep{bernstein2002complexity}, a claim that \acrshort{idreamer} cannot guarantee.

\section{Results}
To evaluate the efficacy of our proposed approach, we select two distinct environment suites that present diverse challenges. Firstly, we utilise \acrshort{vmas} \citep{bettini2022vmas} to examine the performance of our methods with lower dimensional vector-based observations. Secondly, we use Melting Pot \citep{agapiou2022melting} to investigate the efficacy of our methods in higher-dimensional visual observation settings. We evaluate a variety of separate tasks within each environment suite and utilise the evaluation methodology outlined in by \citet{agarwal2021deep}, and further explained in section \ref{eval-method}, to produce aggregated results.

\subsection{Vectorised Multi-Agent Simulator}

\acrfull{vmas}\footnote{Found at https://github.com/proroklab/VectorizedMultiAgentSimulator}\citep{bettini2022vmas} is an open-source, 2D physics simulation platform designed to assess various \acrshort{marl} algorithms across multi-agent coordination problems. \acrshort{vmas} offers diverse, challenging scenarios requiring varying degrees of individual skill and collaboration. The observation spaces are vector-based, simulating real-world robotic sensing systems like LIDAR, and the action spaces can be continuous or discrete. In this work, we evaluate our method on three \acrshort{vmas} scenarios: Flocking, Discovery, and Buzz Wire, all with discrete action spaces.

\begin{figure}[h!]
    \centering
    \includegraphics[width=\linewidth]{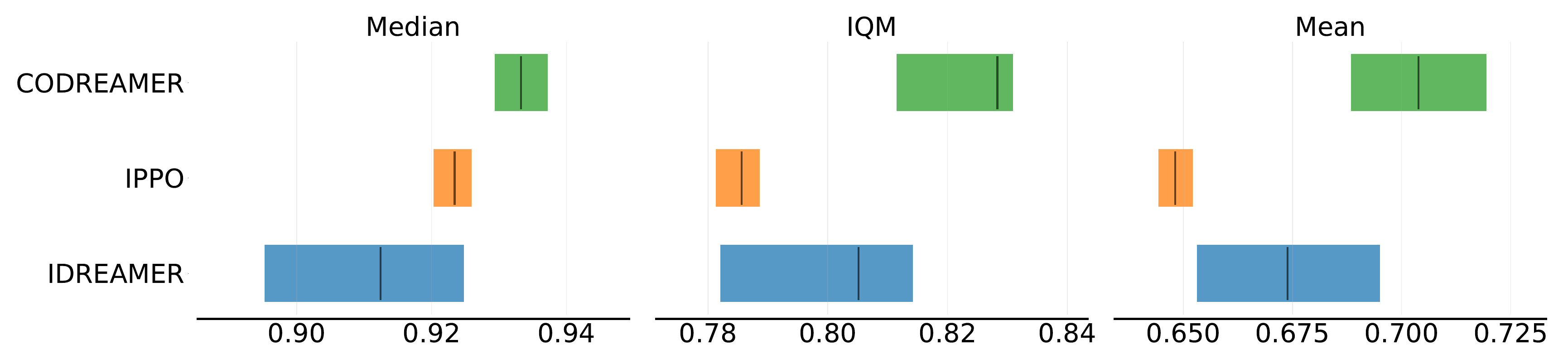}
    \caption[Aggregate metrics on \acrshort{vmas} Tasks]{\textbf{Aggregate metrics on \acrshort{vmas} Tasks} with 95\% \acrshort{ci}s. Reported from left to right are: Median, IQM, Mean.}
    \label{fig:vmas_aggregate}
\end{figure}

In \acrshort{vmas}, \acrshort{codreamer} shows superior performance in all point estimate metrics. Figure \ref{fig:vmas_aggregate} illustrates that while \acrshort{codreamer}'s point estimates surpass both I\acrshort{ppo} and \acrshort{idreamer}, there is some overlap in their \acrshort{iqm} and mean \acrlong{ci}, indicating minor statistical uncertainty. However, the overlap is minor, suggesting with confidence that \acrshort{codreamer} leads to improved results. To understand performance nuances beyond point estimates, we examine additional metrics.

\begin{figure}[h!]
    \centering
    \includegraphics[width=\linewidth]{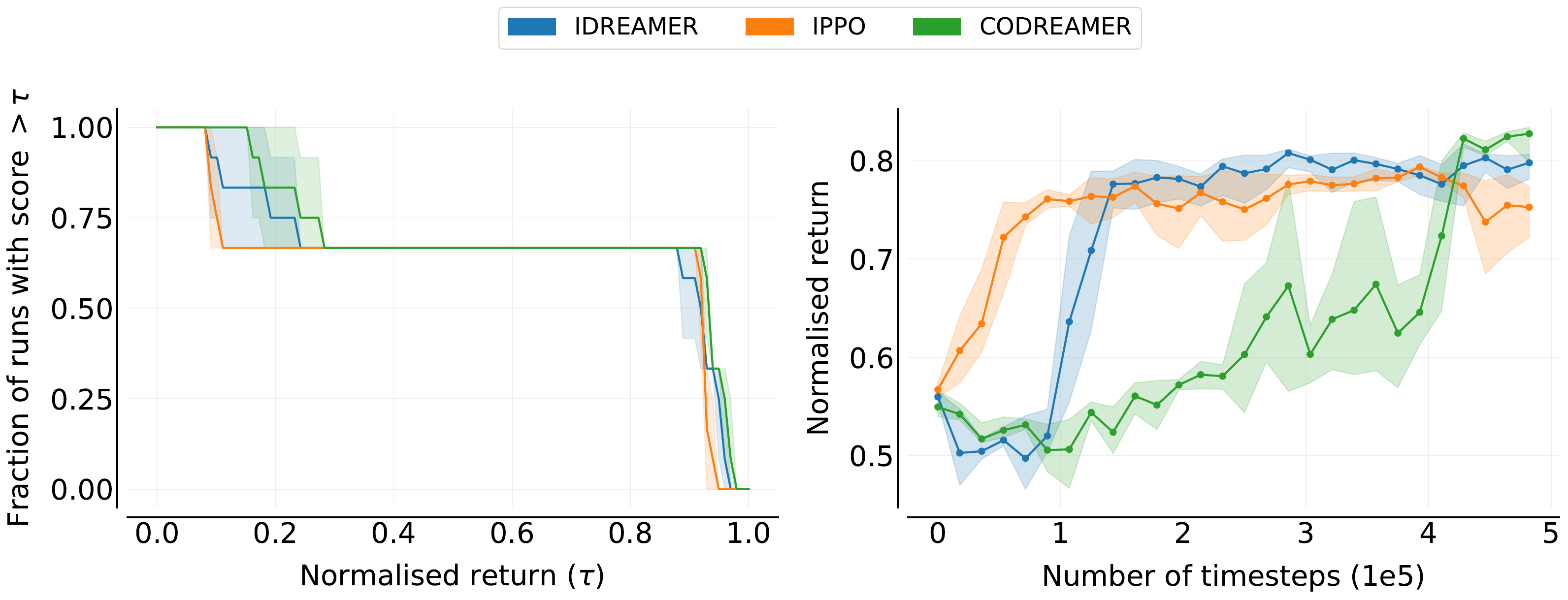}
    \caption[\acrshort{vmas} Evaluation]{\textbf{\acrshort{vmas} Evaluation.} \textbf{Left:} Performance profiles indicating the percentage of runs that scored above a certain normalised return. \textbf{Right:} IQM min-max normalised scores as a function of environment timesteps. This measures the sample efficiency of all agents. For both plots, the shaded regions show 95\% CIs.}
    \label{fig:vmas_perf_sample}
\end{figure}

The performance profile in Figure \ref{fig:vmas_perf_sample} shows that \acrshort{codreamer} consistently matches or exceeds \acrshort{idreamer} and I\acrshort{ppo} across all training runs, suggesting that \acrshort{codreamer}'s communication improves performance. However, the sample efficiency plot reveals a marginal trade-off, with \acrshort{codreamer}'s final performance accompanied by reduced initial sample efficiency. I\acrshort{ppo} and \acrshort{idreamer} reach their final performance with less data. This trade-off is expected, as learning communication protocols adds complexity. Both \acrshort{idreamer} and \acrshort{codreamer} need to learn an accurate environment model before achieving performance gains. While I\acrshort{ppo} is initially more sample efficient, \acrshort{idreamer} quickly surpasses it after 100,000 environment steps.

In conclusion, both \acrshort{codreamer} and \acrshort{idreamer} statistically significantly outperform I\acrshort{ppo}, with \acrshort{codreamer} showing the best results among all algorithms. However, the performance improvements are relatively marginal.

\subsection{Melting Pot}

Melting Pot\footnote{Found at https://github.com/google-deepmind/meltingpot}\citep{agapiou2022melting} benchmarks \acrshort{marl} by evaluating agents' ability to generalise and adapt to unfamiliar environmental and social contexts. It includes tasks that assess various social interactions such as cooperation, competition, and trust. Agents are trained in specific games and assessed in unique test scenarios to evaluate their generalisation.

In this work, we focus on cooperative performance rather than social generalisation. We train and evaluate agents only on the original games, excluding Melting Pot’s unique test scenarios. Melting Pot's observation space is pixel-based and partially observable, with each agent having a distinct field of view. This setup demonstrates our method's capability to model complex, high-dimensional observations and highlights the benefits of communication in prediction and policy learning. We select four scenarios: Daycare, Cooperative Mining, and two variants of Collaborative Cooking. Each scenario is transformed into a \acrshort{dec-pomdp} by using a single global reward, which is the sum of individual rewards.

\begin{figure}[h!]
    \centering
    \includegraphics[width=\linewidth]{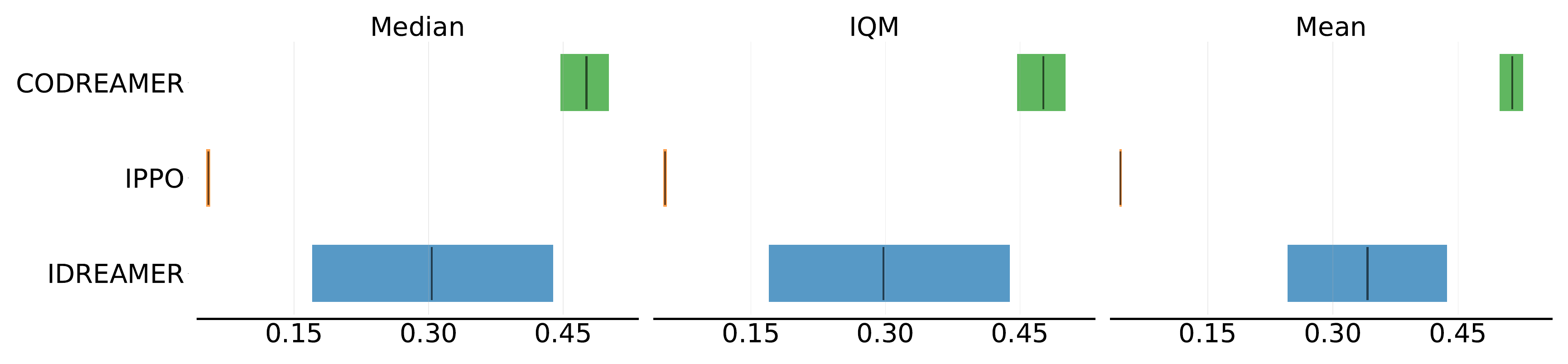}
    \caption[Aggregate metrics on Melting Pot Tasks]{\textbf{Aggregate metrics on Melting Pot Tasks} with 95\% CIs. Reported from left to right are: Median, IQM, Mean.}
    \label{fig:meltingpot_aggregate}
\end{figure}

In Melting Pot tasks, both \acrshort{idreamer} and \acrshort{codreamer} significantly outperform I\acrshort{ppo}, as shown in Figure \ref{fig:meltingpot_aggregate}. I\acrshort{ppo} consistently fails to achieve high scores, which aligns with expectations given the sample inefficiency of pixel-based environments. As a model-free on-policy algorithm, I\acrshort{ppo} struggles with the limited data (500,000 steps) to learn adequate visual features and high-level strategies. In contrast, \acrshort{codreamer} outperforms \acrshort{idreamer} in all aggregate metrics with higher normalised scores and less variance, and the results show no \acrshort{ci} overlap, indicating high certainty.

\begin{figure}[h!]
    \centering
    \includegraphics[width=\linewidth]{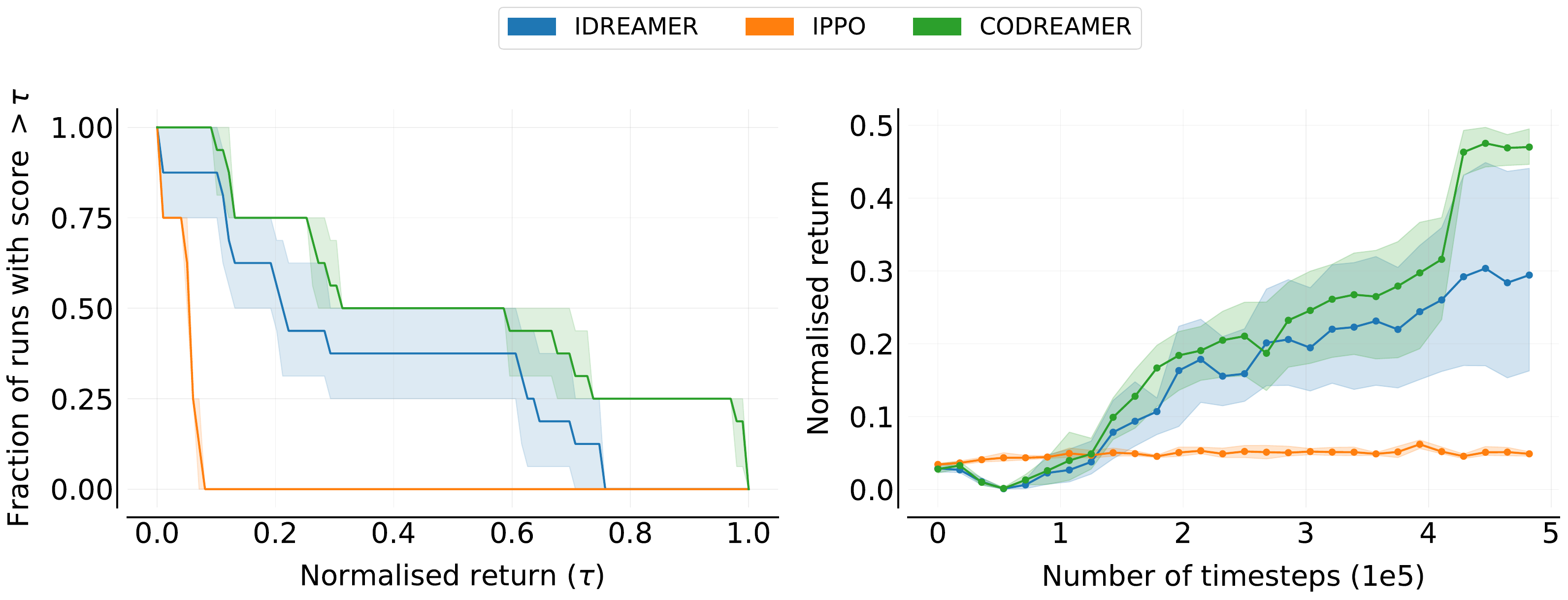}
    \caption[Melting Pot Evaluation]{\textbf{Melting Pot Evaluation.} \textbf{Left:} Performance profiles indicating the percentage of runs that scored above a certain normalised return. \textbf{Right:} IQM min-max normalised scores as a function of environment timesteps. This measures the sample efficiency of all agents. For both plots, the shaded regions show 95\% CIs.}
    \label{fig:coop_perf_sample}
\end{figure}

The performance profile in Figure \ref{fig:coop_perf_sample} shows that \acrshort{idreamer} and \acrshort{codreamer} significantly outperform I\acrshort{ppo}, with I\acrshort{ppo} failing to score above 0.1 in any task or run. \acrshort{codreamer} consistently outperforms \acrshort{idreamer} with low \acrshort{ci} overlap, indicating stochastic dominance. Additionally, \acrshort{codreamer} achieves near-normalised scores of 1.0 in 25\% of runs, showing its capability to consistently achieve maximum rewards. Despite the added complexity of communication, \acrshort{codreamer} demonstrates higher performance with less data, proving the benefit of increased expressivity in learning.

\section{Related Work}

\subsection{\acrshort{marl} With Learned Communication} 
End-to-end learning of communication protocols has been used to create efficient communication mechanisms without needing significant domain expertise \citep{sukhbaatar2016learning, foerster2016learning, peng2017multiagent, kong2017revisiting, jiang2018learning}.

\acrshort{gnn}s have become popular for facilitating communication between agents in various \acrshort{marl} tasks, including cooperative navigation, traffic control, and robotic swarms \citep{li2020graph, tomoki2018, tolstaya2020learning}. These methods efficiently propagate relevant information without breaking the decentralised nature of the system and handle dynamic and heterogeneous environments \citep{bettini2023heterogeneous, li2020graph}. However, traditional \acrshort{gnn}s have not been used within world models for model-based \acrshort{marl}, setting the stage for \acrshort{codreamer}.

\subsection{Multi-Agent Model-Based \acrshort{rl} Methods}

\textbf{Dyna-style} methods, like \acrfull{mambpo} \citep{willemsen2021mambpo} and \acrfull{aorpo} \citep{zhang2021model}, use data from both the real environment and learned models to train agent policies. While some approaches benefit from centralisation, \acrshort{codreamer} offers a fully decentralised framework. Unlike \acrshort{aorpo}, \acrshort{codreamer} does not need to explicitly model all other agents, as each agent's world model implicitly captures information about others.

\textbf{Direct-style} methods, such as \acrfull{mamba} \citep{egorov2022scalable}, learn models from environment data and update agent policies accordingly. \acrshort{codreamer} incorporates a two-level communication system, utilising \acrshort{gnn}s for graph-like structures, and integrates communication into the world models' computation. \acrshort{mamba} extends DreamerV2 \citep{hafner2021mastering} with alterations such as using MA\acrshort{ppo} to train policies.

\textbf{Communication-based} methods, like \acrfull{intshare} \citep{kim2021communication} and \acrfull{maci} \citep{pretoriuslearning}, use environment models to convey agents' long-term future predictions. While \acrshort{codreamer} does not specifically use a world model for this purpose, it effectively facilitates communication to create the agents' current state representation for each timestep.

\section{Conclusion}
In this work, we tackle the issue of performance given a limited sample budget in \acrshort{marl} by exploring model-based \acrshort{rl} as a solution. We adapt the single-agent DreamerV3 algorithm \citep{hafner2023mastering} into an independent multi-agent variant, \acrshort{idreamer}. Recognizing limitations like partial observability, non-stationarity, and inter-agent dynamics, we introduce \acrshort{codreamer}, which enhances \acrshort{idreamer} with a two-level \acrshort{gnn}-based communication system to improve global observability, handle non-stationarity, and model inter-agent dependencies.

Our evaluations show statistically significant improvements of \acrshort{codreamer} over \acrshort{idreamer} and the well-known \acrshort{marl} algorithm I\acrshort{ppo}, especially in environments with high-dimensional visual observations. These results demonstrate the potential of \acrshort{codreamer} and similar model-based \acrshort{marl} methods for sample-efficient learning. The issues it addresses are relevant to real-world applications like multi-robot systems and on-robot learning. We see \acrshort{codreamer} as a promising step toward scalable, communicative model-based \acrshort{marl}, with potential for significant real-world impact.

%%%%%%%%%%%%%%%%%%%%%%%%%%%%%%%%%%%%%%%%%%%%%%%%%%%%%%%%%%%%%%%%
%% NOTE: THIS MARKS THE END OF THE "MAIN TEXT"
%%%%%%%%%%%%%%%%%%%%%%%%%%%%%%%%%%%%%%%%%%%%%%%%%%%%%%%%%%%%%%%%

%%%%%%%%%%%%%%%%%%%%%%%%%%%%%%%%%%%%%%%%%%%%%%%%%%%%%%%%%%%%%%%%
%% Bibliography
%%%%%%%%%%%%%%%%%%%%%%%%%%%%%%%%%%%%%%%%%%%%%%%%%%%%%%%%%%%%%%%%
\bibliography{main}
\bibliographystyle{rlc}

%%%%%%%%%%%%%%%%%%%%%%%%%%%%%%%%%%%%%%%%%%%%%%%%%%%%%%%%%%%%%%%%
%% Appendices
%%%%%%%%%%%%%%%%%%%%%%%%%%%%%%%%%%%%%%%%%%%%%%%%%%%%%%%%%%%%%%%%
\appendix
\section{Additional Results}

\begin{figure}[h!]
    \centering
    \includegraphics[width=0.9\linewidth]{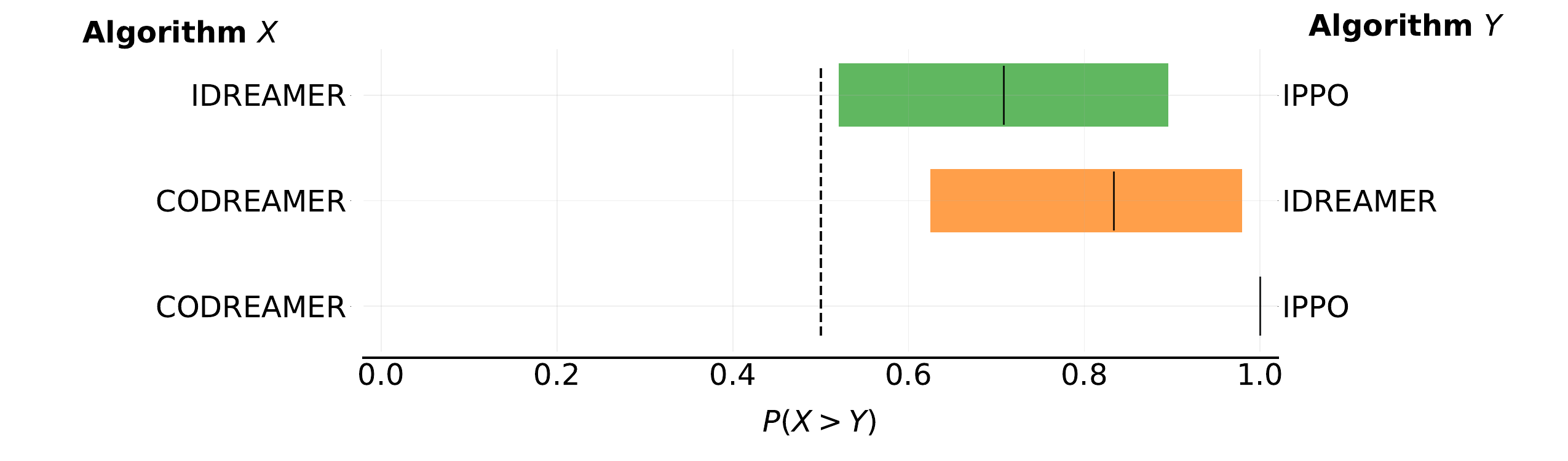}
    \caption[Probability of Improvement on \acrshort{vmas} Tasks]{\textbf{Probability of Improvement on \acrshort{vmas} Tasks.} Each row shows the probability of improvement, with 95\% CIs, that algorithm $X$ outperforms algorithm $Y$.}
    \label{fig:vmas_prob_imp}
\end{figure}

Figure \ref{fig:vmas_prob_imp} shows that \acrshort{codreamer} has a 100\% probability of improvement over I\acrshort{ppo} and over 80\% probability of outperforming \acrshort{idreamer}. \acrshort{idreamer} also has a high probability of outperforming IPPO. Since all lower bounds of \acrshort{ci}s are above 0.5 and the upper bounds are above 0.75, the results are statistically significant and meaningful \citep{agarwal2021deep}.

\begin{figure}[h!]
    \centering
    \includegraphics[width=0.9\linewidth]{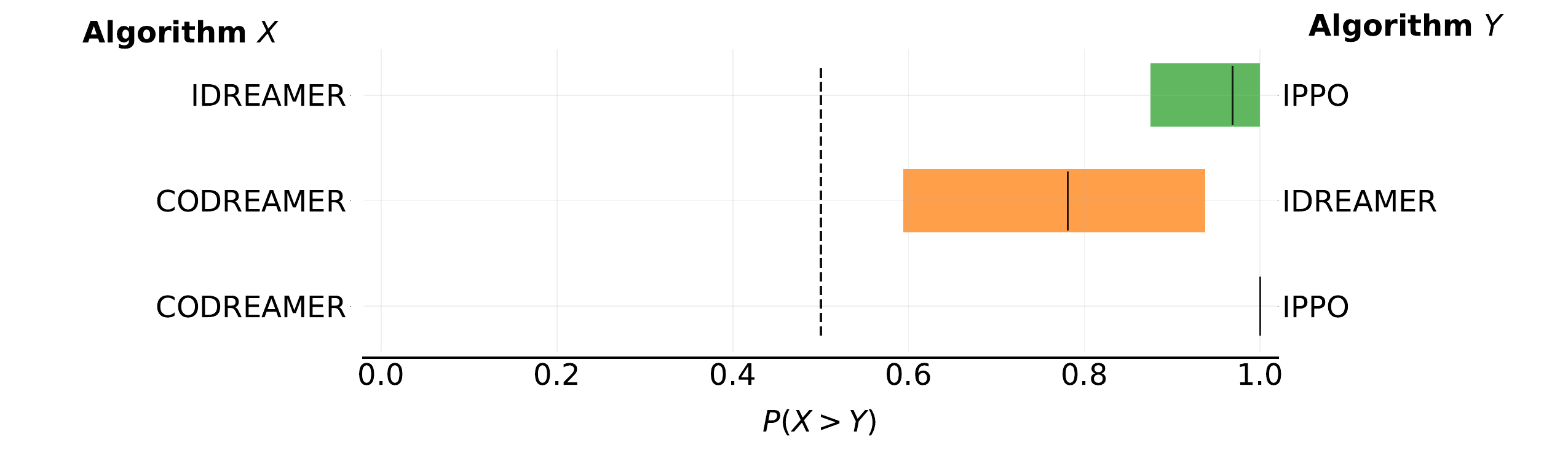}
    \caption[Probability of Improvement on Melting Pot Tasks]{\textbf{Probability of Improvement on Melting Pot Tasks.} Each row shows the probability of improvement, with 95\% CIs, that algorithm $X$ outperforms algorithm $Y$.}
    \label{fig:coop_prob_imp}
\end{figure}

Figure \ref{fig:coop_prob_imp} shows similar results in the probability of improvement metrics. \acrshort{idreamer} has a high probability of improvement over I\acrshort{ppo}, and \acrshort{codreamer} has a 100\% probability of outperforming I\acrshort{ppo}. \acrshort{codreamer} also has a higher probability of outperforming \acrshort{idreamer} in Melting Pot tasks. Although \acrshort{codreamer}'s performance improvement is larger in magnitude in Melting Pot compared to \acrshort{vmas}, the lower probability of improvement and \acrshort{ci} indicates slightly greater uncertainty. However, all probabilities are statistically significant and meaningful.

\section{Empirical Validation of Modelling Expressivity}

To empirically validate our expressivity claims, we present a modified version of the Estimate Game \citep{kortvelesy2022qgnn} known as the Sequential Estimate Game. This environment extends the original game into a multi-step \acrfull{pomg}, which cannot be accurately represented solely based on the information of individual agents.

Commonly used environments \citep{samvelyan19smac,ellis2022smacv2, kurach2020google, papoudakis2021benchmarking} in \acrshort{marl} literature suggest that cooperation and coordination are required to achieve an optimal policy. However, this is often simply conjectured and generally remains unknown. Due to this fact, the original variant of the Estimate Game was proposed to serve as a stress test for multi-agent cooperative systems. The environment is inherently simple yet provably requires coordination through communication to solve by explicitly constructing a dependency for agents to utilise non-local information.

In the original Estimate Game, all $n$ agents are assigned a local state $s^i \in [0, 1]$ where each agent's observation is directly their state $o^i = s^i$. Concurrently, a random adjacency matrix $\mathbf{A}$ is constructed, using an edge density $\rho$, that represents agent connectivity and defines which agents are interdependent. 

The explicit interdependence on non-local information is constructed through the reward function which is defined as follows for agent $i$ at timestep $t$:

\[ y^i_t = 2\cdot\Big(\eta\cdot(s^i_t - 0.5) + (1-\eta)\cdot\frac{1}{\mathcal{N}^i}\sum_{j\in\mathcal{N}^i}(s^j_t - 0.5)\Big) + 0.5 \]
\[ r^i_t = - max(|\frac{a^i_t}{|\mathcal{A}|} + \frac{1}{2\cdot|\mathcal{A}|} - y^i_t|) - \frac{1}{2\cdot|\mathcal{A}|}, 0) \]
where $\mathcal{N}^i$ is agent $i$'s neighbours defined by $\mathbf{A}$. 

In simpler terms, this reward function defines the task of agent $i$ predicting the constructed target $y^i$ which is dependent on its neighbours $\mathcal{N}^i$. To make the action space discrete, each agent has a predetermined number of actions $|\mathcal{A}|$ that correspond to equally spaced intervals between 0 and 1. The original Estimate game utilises a global reward to be represented as a \acrshort{dec-mdp} which is the minimum of all local rewards. However, we train agents using their local rewards in order to thoroughly evaluate \acrshort{codreamer}'s ability to learn more independent reward functions that are still dependent on non-local information. Thus, the Sequential Estimate Game can be viewed as a \acrshort{pomg} with agent observations being equal to agent local states.

Our extension to the Estimate Game involves the construction of a simple transition function:

\[s^i_{t+1} = \frac{1}{2}\cdot\cos(a^i_t + \eta*s^i_t + (1-\eta)\cdot\frac{1}{\mathcal{N^i}}\sum_{j\in\mathcal{N}^i}s^j_t) + \frac{1}{2}\]

This transition function allows agents to retain unique states without the risk of all agents converging to the same value due to neighbourhood aggregation. Additionally, as it is constructed similarly to the reward function, it introduces an inter-agent dependence. In the sequential variant of the Estimate Game, the randomly generated adjacency matrix $\mathbf{A}$ is held constant for the entirety of an episode thus giving certain agents complete independence in their reward and transition functions. The use of a static adjacency matrix per episode is done to evaluate each method's ability to both model independent agents and inter-dependent agents simultaneously. We list the specific values used in our implementation of the Sequential Estimate Game in Table \ref{tab:estimategamespecifics}.

\begin{table}[h!]
\centering
\begin{tabular}{@{}lcc@{}}
\toprule
\textbf{Name}                   & \textbf{Symbol}           & \textbf{Value}   \\ \midrule
Number of Agents       & $n$              & 4       \\
Number of Timesteps    & $T$              & 5       \\
Observation Size       & $|s^i_t|$        & 1       \\
Initial State          & $s^i_0$          & U(0, 1) \\
Number of Actions      & $| \mathcal{A}|$ & 4       \\
Edge Density           & $\rho$           & 0.6     \\
Local State Percentage & $\eta$           & 0.3     \\ \bottomrule
\end{tabular}
\caption{Specific instantiation of the Sequential Estimate Game evaluated}
\label{tab:estimategamespecifics}
\end{table}

\label{estimate-game-results}

\begin{figure}[h!]
    \centering
    \includegraphics[width=\linewidth]{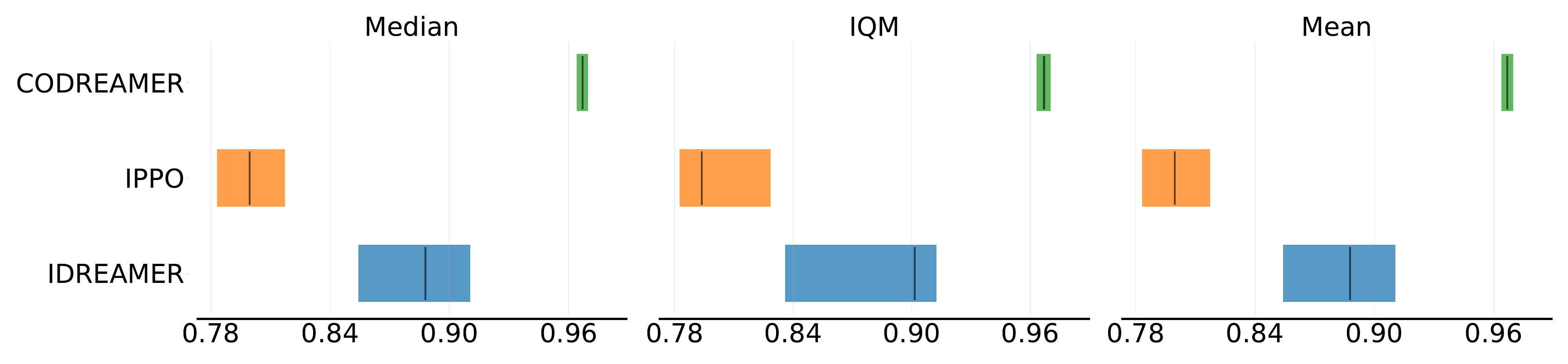}
    \caption[Aggregate metrics on Sequential Estimate Game]{\textbf{Aggregate metrics on Sequential Estimate Game} with 95\% CIs. Reported from left to right are: Median $(\uparrow)$, IQM $(\uparrow)$, Mean $(\uparrow)$. We use the $(\uparrow, \downarrow)$ notation to indicate whether higher or lower scores are desired.}
    \label{fig:estimate_aggregate}
\end{figure}

We present the final aggregated results in Figure \ref{fig:estimate_aggregate}. As expected, \acrshort{codreamer} achieves near-optimal performance with very low variance between runs. Notably, \acrshort{idreamer} outperforms I\acrshort{ppo}. We hypothesise that due to \acrshort{idreamer} learning the dynamics for all agents, it is likely aware of the episodes where an agent is independent thus allowing it to learn policies specifically for these agents when possible.

\begin{figure}[h!]
    \centering
    \includegraphics[width=\linewidth]{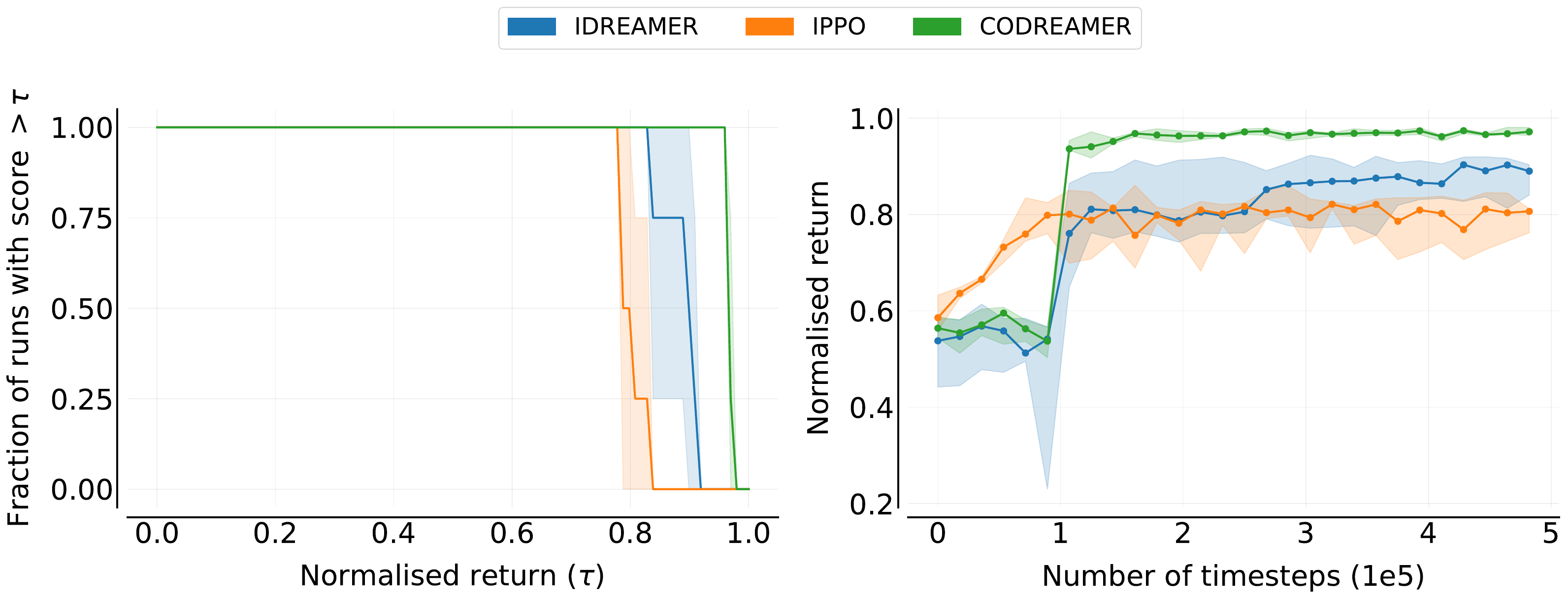}
    \caption[Sequential Estimate Game Evaluation]{\textbf{Sequential Estimate Game Evaluation.} \textbf{Left:} Performance profiles indicating the percentage of runs that scored above a certain normalised return. \textbf{Right.} IQM min-max normalised scores as a function of environment timesteps. This measures the sample efficiency of all the agents. For both plots, the shaded regions show 95\% CIs.}
    \label{fig:estimate_perf_sample}
\end{figure}

Figure \ref{fig:estimate_perf_sample} shows the performance profile and sample efficiency of each algorithm. The performance profile confirms our aggregated metrics and shows that \acrshort{codreamer} has very low variance in its performance over all runs. Additionally, we see that all methods converge to their final performance at around 100,000 - 200,000 timesteps. Interestingly, we see that \acrshort{idreamer} achieves a significant increase in performance at around 300,000 timesteps. This further gives evidence to our hypothesis that, given enough data, \acrshort{idreamer} manages to learn the optimal policy for independent agents specifically. It is possible that through the use of a recurrent I\acrshort{ppo} baseline, the same policy would be found.

\begin{figure}[h!]
    \centering
    \includegraphics[width=\linewidth]{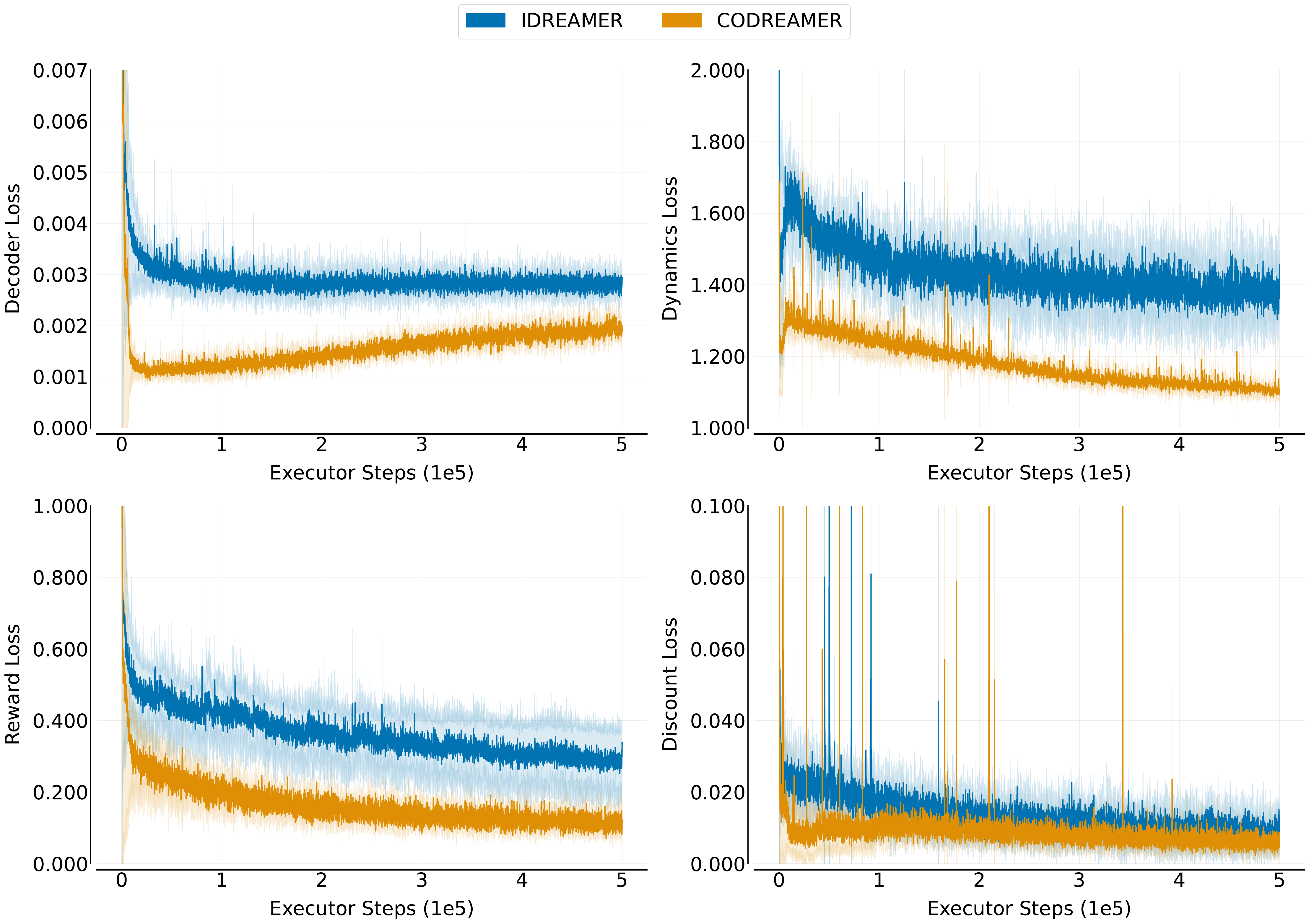}
    \caption[Mean prediction loss curves]{\textbf{Mean prediction loss curves} over all Sequential Estimate Game training runs. \acrshort{codreamer} significantly achieves lower loss values in predicting each respective environment quantity.}
    \label{fig:estimate_wm_losses}
\end{figure}

Although not conventional, we present the relevant loss curves in Figure \ref{fig:estimate_wm_losses} to further evaluate the world modelling differences between \acrshort{codreamer} and \acrshort{idreamer}. In the context of the Sequential Estimate Game, an environment crafted to explicitly have a high degree of inter-agent dependence, we see that \acrshort{codreamer} achieves much lower losses over the course of training. As each loss directly correlates to prediction accuracy, it is fair to claim that the modelling of the environment is much more accurate using \acrshort{codreamer}.

Although these results confirm our initial claims, we acknowledge that the Sequential Estimate Game is a highly limited environment that might not be representative of more realistic \acrshort{marl} use cases.

\section{Ablation}

Following our initial set of experiments, we sought to isolate the individual impact of each distinct level of communication within \acrshort{codreamer}. Consequently, we evaluate a separate set of experiments where only a single level of communication is utilised either within the world model or actor-critic. Particularly, when limiting \acrshort{codreamer} to use communication exclusively within the world model, our objective is to investigate if the learnt state representation holds enough information for independent actor-critic networks to achieve high performance. For the ablation results, we term the separate levels of communication as WM Comm for the world model and AC comm for the actor-critic.

\subsection{Estimate Game}

\begin{figure}[h!]
    \centering
    \includegraphics[width=\linewidth]{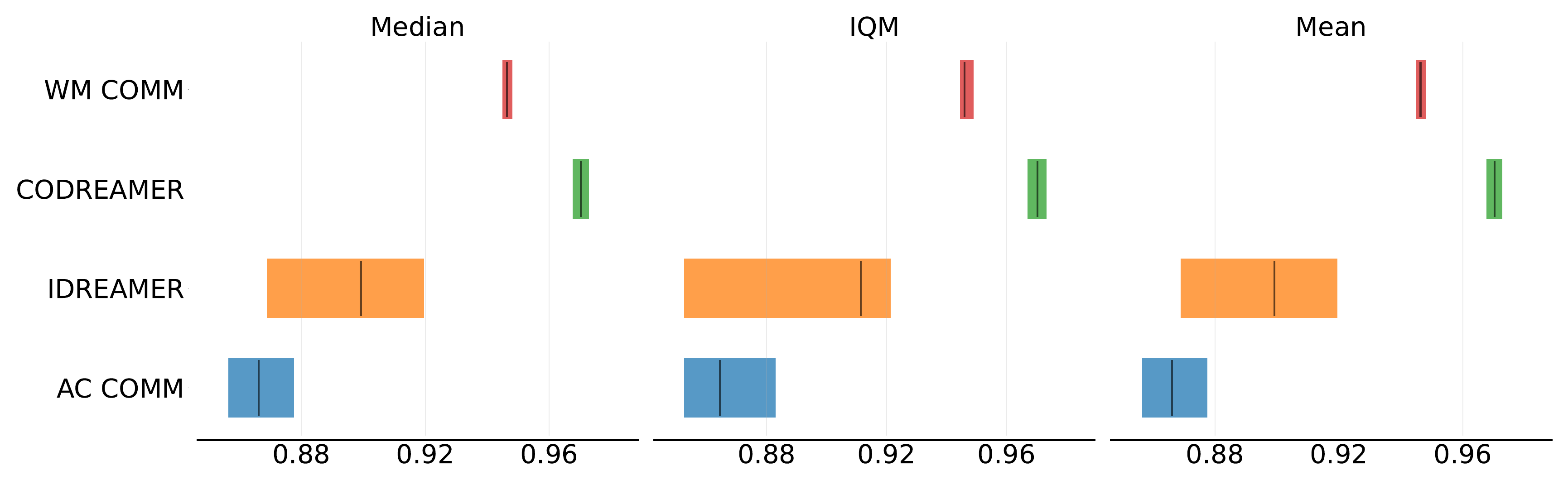}
    \caption[Aggregate metrics for Sequential Estimate Game Ablation]{\textbf{Aggregate metrics for Sequential Estimate Game Ablation} with 95\% CIs. Reported from left to right are: Median $(\uparrow)$, IQM $(\uparrow)$, Mean $(\uparrow)$. We use the $(\uparrow, \downarrow)$ notation to indicate whether higher or lower scores are desired.}
    \label{fig:ablation-estimate-aggregate}
\end{figure}

Interestingly, we observe in Figure \ref{fig:ablation-estimate-aggregate} that \acrshort{idreamer}, which utilises no communication, outperforms AC Comm. This is seemingly due to how Direct-style model-based methods train. If the world model is unable to capture the dynamics and reward functions of the environment, then the learnt representations of the state will be largely incorrect. As this is the case, using communication on top of the inaccurate state representations as well as generating and training on synthetic data will lead to sub-optimal policies. This would naturally make one assume that the performance of \acrshort{idreamer} and AC Comm would perform the same, however, the added complexity of communication most likely inhibits learning from even performing well on more independent agents. Remarkably, we see that utilising communication in the world model exclusively i.e. WM Comm, allows the independent actor-critic networks to perform significantly better than \acrshort{idreamer}. This indicates that there can be a large degree of overlap in the information that is relevant for the world model and that which is relevant for the actor-critic networks. However, we do note that in the particular instance of the Sequential Estimate Game, this overlap of information is highly evident by the construction of the environment.

\begin{figure}[h!]
    \centering
    \includegraphics[width=\linewidth]{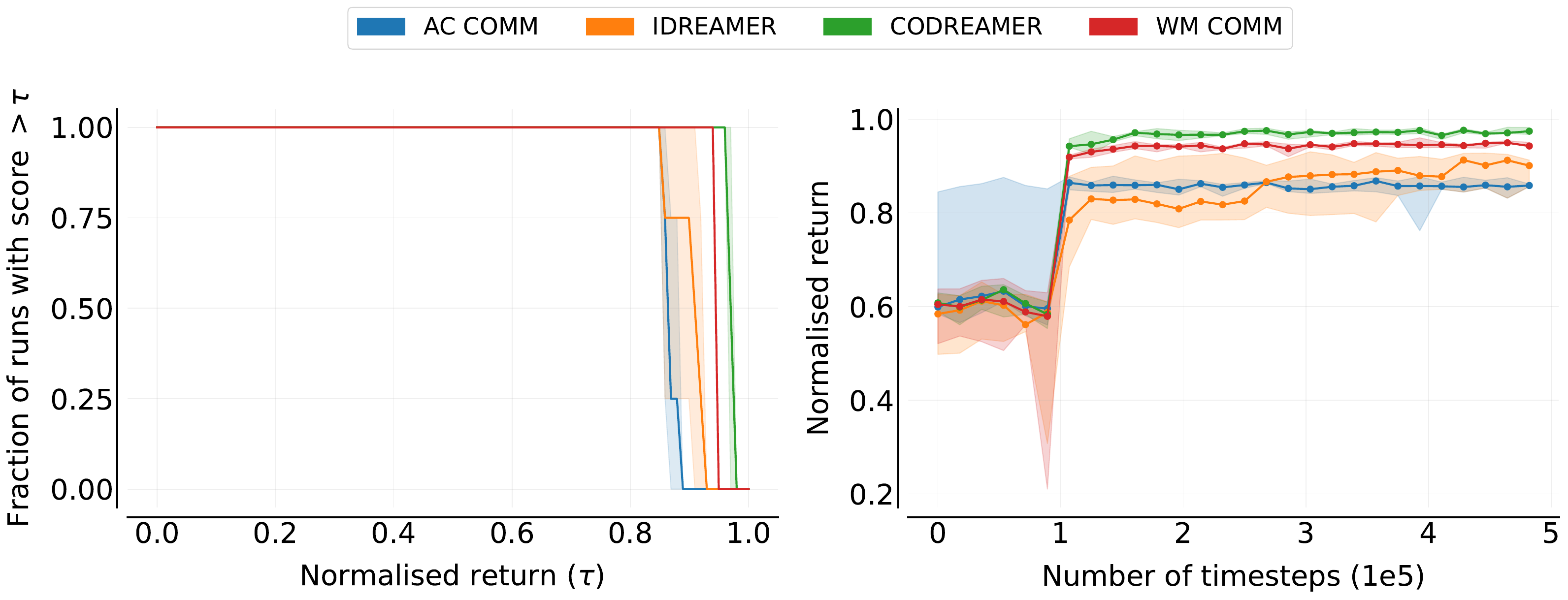}
    \caption[Sequential Estimate Game Ablation]{\textbf{Sequential Estimate Game Ablation.} \textbf{Left:} Performance profiles indicating the percentage of runs that scored above a certain normalised return. \textbf{Right.} IQM min-max normalised scores as a function of environment timesteps. This measures the sample efficiency of all the agents. For both plots, the shaded regions show 95\% CIs.}
    \label{fig:ablation-estimate-perf-sample}
\end{figure}

The performance profiles and sample efficiency plots in Figure \ref{fig:ablation-estimate-perf-sample} give results consistent with those in Section \ref{estimate-game-results}. We see that all methods converge quickly to their final performance. Both \acrshort{idreamer} and AC Comm have a much higher degree of variance in their performance compared to \acrshort{codreamer} and WM Comm. This further confirms our hypothesis that both \acrshort{idreamer} and AC Comm obtain their performance through modelling the agents that are seemingly independent, which change every episode thereby introducing potential high variance in returns.

\subsection{VMAS}

\begin{figure}[h!]
    \centering
    \includegraphics[width=\linewidth]{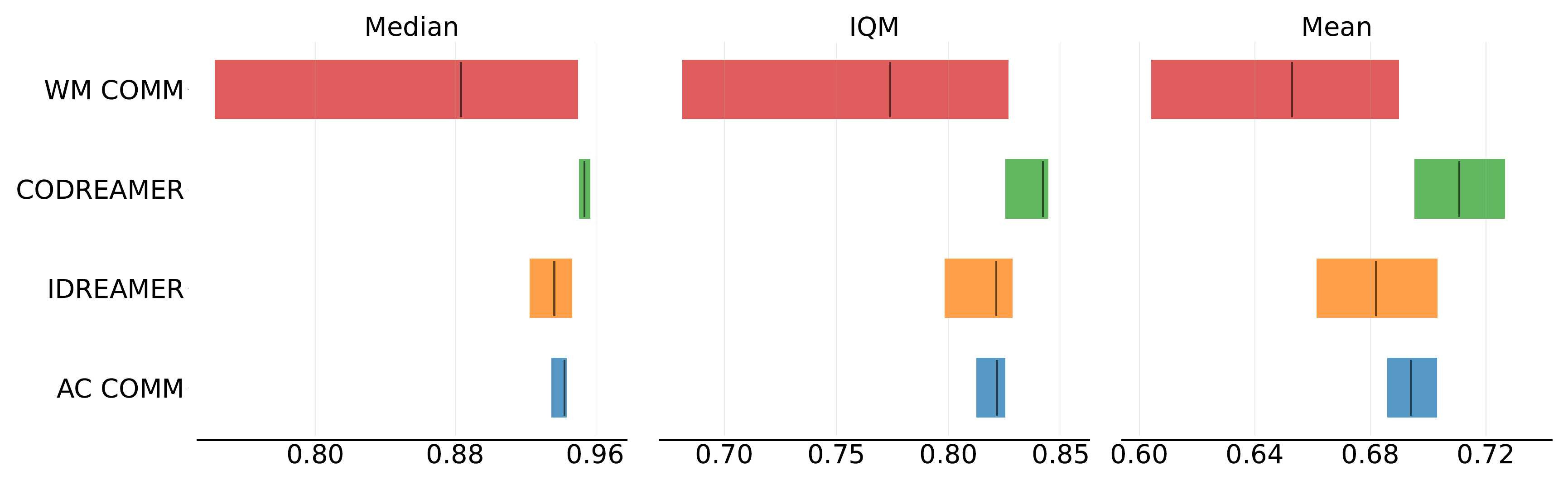}
    \caption[Aggregate metrics for VMAS Ablation]{\textbf{Aggregate metrics for VMAS Ablation} with 95\% CIs. Reported from left to right are: Median $(\uparrow)$, IQM $(\uparrow)$, Mean $(\uparrow)$. We use the $(\uparrow, \downarrow)$ notation to indicate whether higher or lower scores are desired.}
    \label{fig:ablation-vmas-aggregate}
\end{figure}

Looking at the aggregate metrics in Figure \ref{fig:ablation-vmas-aggregate}, we see that when exclusively utilising communications within the world model, performance significantly decreases. These results are counter-intuitive as we would expect that a similar pattern to the estimate game would emerge where a more shared state representation could assist the independent actor-critic networks. This seems to indicate that the evaluated VMAS tasks' reward and transition functions are not highly inter-agent dependent thereby not benefitting from the introduced expressivity. However, we do see that when adding communication in the actor-critic networks, performance gains are had yet the results are still close in magnitude to \acrshort{idreamer}. As noted before, \acrshort{codreamer} outperforms \acrshort{idreamer} which has no communication suggesting that the combination of both levels of communication can yield a unique improvement that no individual part can give. Ultimately, we see from the confidence intervals that \acrshort{idreamer} and \acrshort{codreamer} are similar in performance and improvements can be due to experiment stochasticity.

\begin{figure}[h!]
    \centering
    \includegraphics[width=\linewidth]{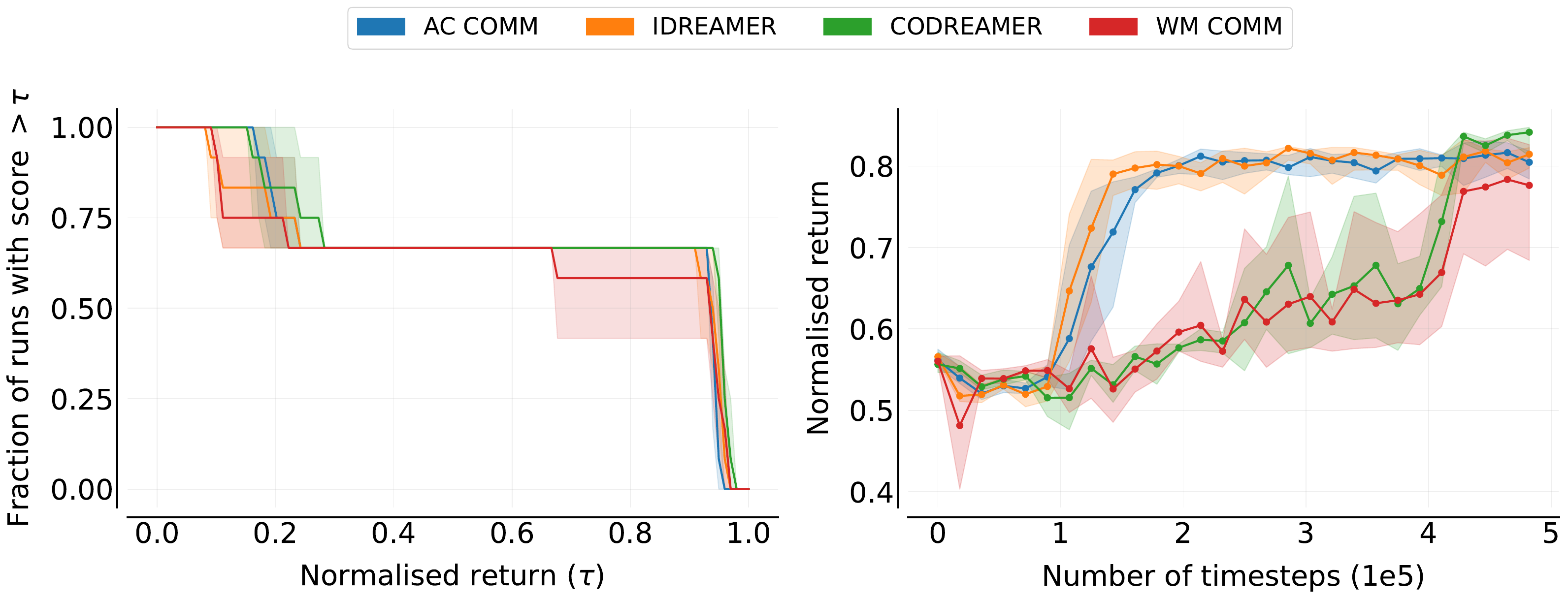}
    \caption[VMAS Ablation]{\textbf{VMAS Ablation.} \textbf{Left:} Performance profiles indicating the percentage of runs that scored above a certain normalised return. \textbf{Right.} IQM min-max normalised scores as a function of environment timesteps. This measures the sample efficiency of all the agents. For both plots, the shaded regions show 95\% CIs.}
    \label{fig:ablation-vmas-perf-sample}
\end{figure}

The sample efficiency plots presented in Figure \ref{fig:ablation-estimate-perf-sample} demonstrate that \acrshort{idreamer} and AC Comm exhibit similar levels of sample efficiency. We see the same observation for \acrshort{codreamer} and WM Comm. These findings suggest that communication within the world model acts as the limiting factor in sample efficiency, as increased expressivity negatively impacts early performance during the training process.

\begin{figure}[h!]
\centering
    \includegraphics[width=\linewidth]{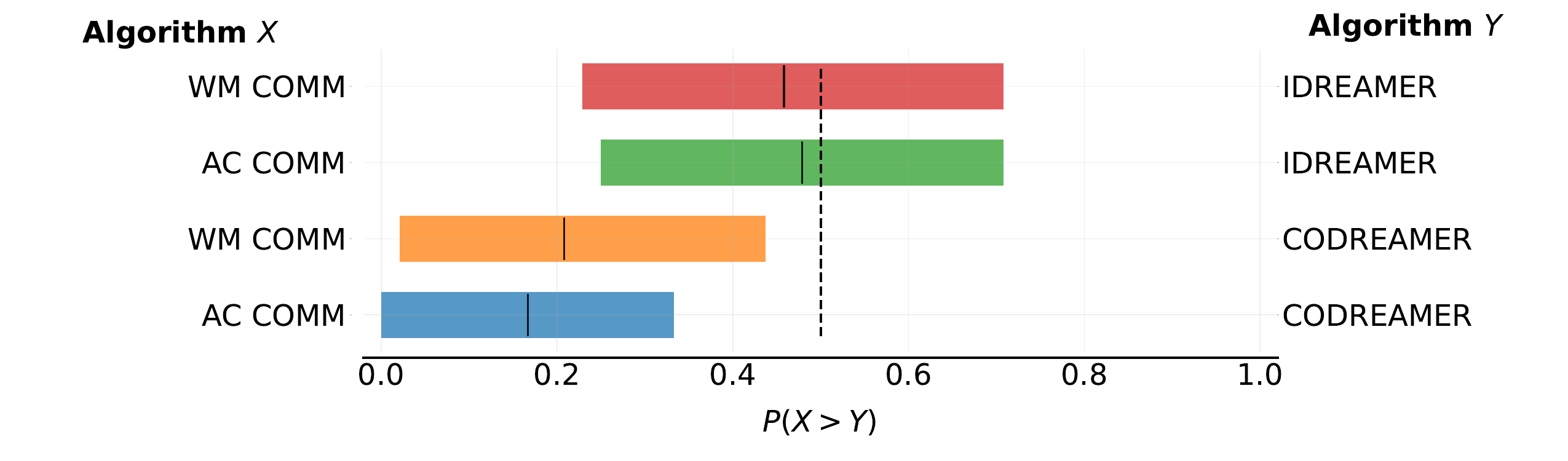}
    \caption[Probability of Improvement for VMAS Ablation]{\textbf{Probability of Improvement for VMAS Ablation.} Each row shows the probability of improvement, with 95\% CIs, that algorithm $X$ outperforms algorithm $Y$.}
    \label{fig:ablation-vmas-prob}
\end{figure}

In Figure \ref{fig:ablation-vmas-prob}, we observe that both AC Comm and WM Comm have a lower probability of improvement compared to \acrshort{idreamer}. However, these results lack statistical significance or meaning as the \acrshort{ci}s lower and upper bound do not surpass 0.5 and 0.75 respectively. Thus, we cannot say with certainty that given a random task that IDreamer will perform better. Additionally, we see that both communication levels exhibit a lower probability of improvement compared to \acrshort{codreamer}, but these differences are statistically significant and meaningful.

\subsection{Melting Pot}

\begin{figure}[h!]
    \centering
    \includegraphics[width=\linewidth]{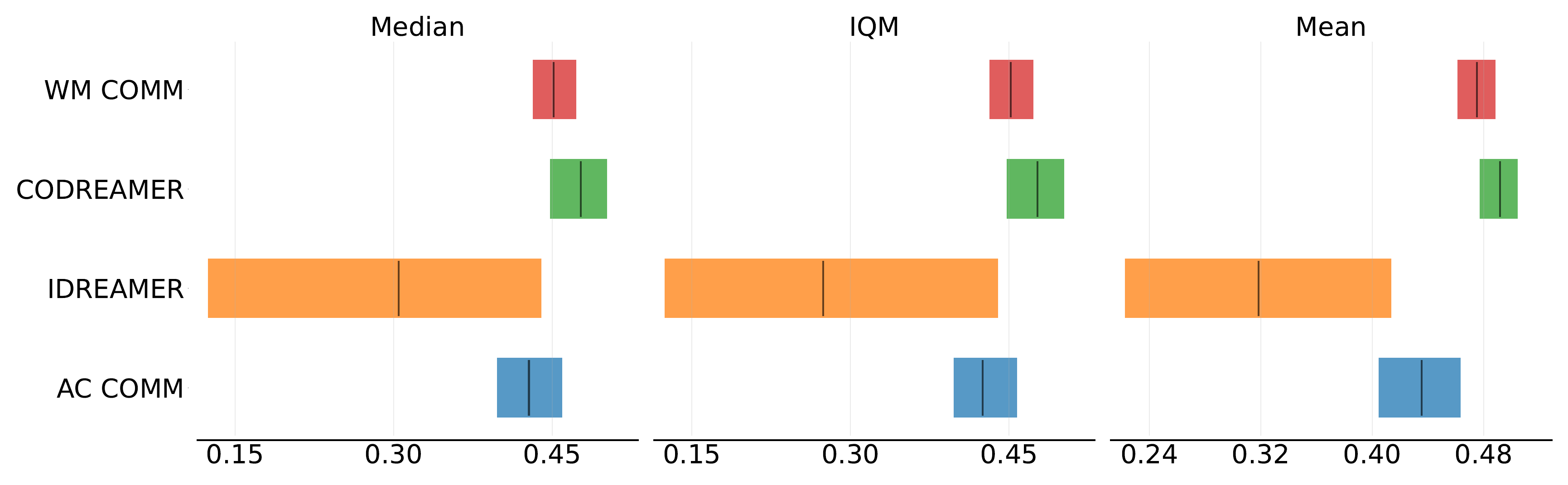}
    \caption[Aggregate metrics for Melting Pot Ablation]{\textbf{Aggregate metrics for Melting Pot Ablation} with 95\% CIs. Reported from left to right are: Median $(\uparrow)$, IQM $(\uparrow)$, Mean $(\uparrow)$. We use the $(\uparrow, \downarrow)$ notation to indicate whether higher or lower scores are desired.}
    \label{fig:ablation-meltingpot-aggregate}
\end{figure}

We see when looking at the results presented in Figure \ref{fig:ablation-meltingpot-aggregate}, that \acrshort{codreamer} outperforms all other methods in every point estimate. This indicates that the combination of both levels of communication provides consistent improvements across all tasks in Melting Pot, with no specific outliers. Interestingly, closely following \acrshort{codreamer}'s performance is WM comm and then AC Comm informing us that communication in the world model provided the largest performance gains. WM Comm's \acrshort{ci}'s have a large overlap with \acrshort{codreamer} indicating that there is uncertainty in the improvements provided by the second level of communication. This seems to suggest that the state representation learnt is beneficial and provides enough shared information for the independent actor-critic networks to exploit and cooperate without needing their own form of explicit communication. We posit that the Melting Pot environments, due to their visual nature, give each agent's state representation enough information for an independent actor-critic network to know what their teammates are planning on doing, thus simply utilising communications in the world model is enough to improve cooperation.

\begin{figure}[h!]
    \centering
    \includegraphics[width=\linewidth]{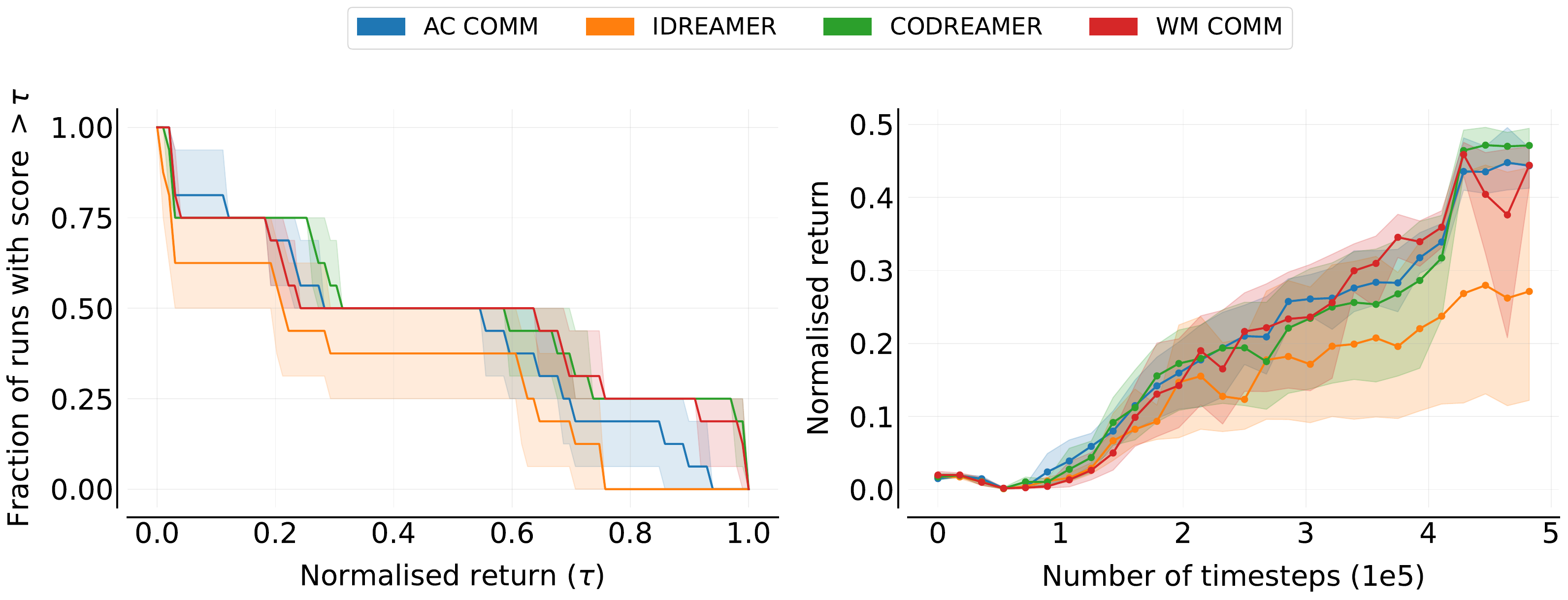}
    \caption[Melting Pot Ablation]{\textbf{Melting Pot Ablation.} \textbf{Left:} Performance profiles indicating the percentage of runs that scored above a certain normalised return. \textbf{Right.} \acrshort{iqm} min-max normalised scores as a function of environment timesteps. This measures the sample efficiency of all the agents. For both plots, the shaded regions show 95\% CIs.}
    \label{fig:ablation-meltingpot-perf-sample}
\end{figure}

The performance profiles in Figure \ref{fig:ablation-estimate-perf-sample} show us that although AC Comm has a slightly higher lower bound in performance, both \acrshort{codreamer} and WM Comm have higher upper bounds indicating that certain environments do greatly benefit from the communicative world models whereas other environments can be negatively impacted due to the increased complexity albeit marginally. Furthermore, we observe stochastic dominance of \acrshort{idreamer} by all communicative methods. Lastly, We observe in the sample efficiency plots that all communicative methods experience roughly the same sample efficiency with no method being clearly superior. 

\begin{figure}[h!]
\centering
    \includegraphics[width=\linewidth]{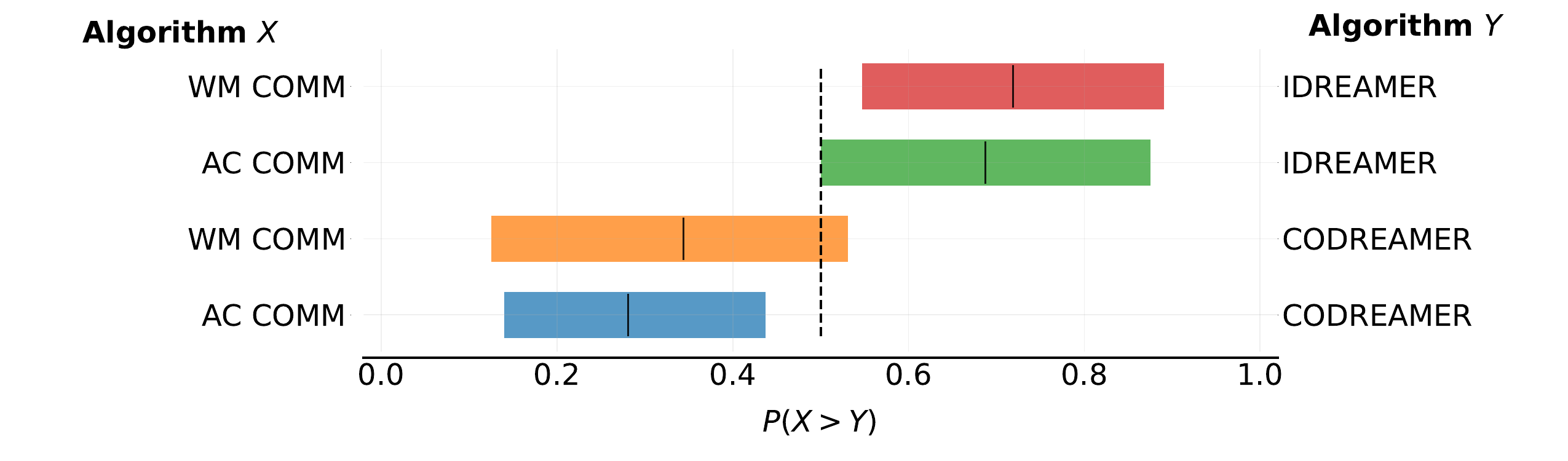}
    \caption[Probability of Improvement for Melting Pot Ablation]{\textbf{Probability of Improvement for Melting Pot Ablation.} Each row shows the probability of improvement, with 95\% CIs, that algorithm $X$ outperforms algorithm $Y$.}
    \label{fig:ablation-meltingpot-prob}
\end{figure}

In the probability of improvement, presented in Figure \ref{fig:ablation-meltingpot-prob}, we see interesting results. Unlike in \acrshort{vmas}, both AC Comm and WM Comm have a probability greater than 0.5 of improving upon \acrshort{idreamer} performance with both methods achieving statistical significance and meaning. When compared to \acrshort{codreamer}, we see the opposite relationship with both AC Comm and WM Comm being below the 0.5 value. However, in this instance, we see that WM Comm's result is not statistically significant. What we can interpret from these results is that, regardless of the magnitude of improvement, in Melting Pot tasks, communication in either level of \acrshort{codreamer} offers improvements, however, we cannot say with statistical certainty that the use of communication within the actor-critic networks, in addition to the world model, improved performance as CoDreamer and WM Comm perform similarly. 

\section{Experimental Details}

\begin{figure}[h!]
    \centering
    \includegraphics[width=0.75\linewidth]{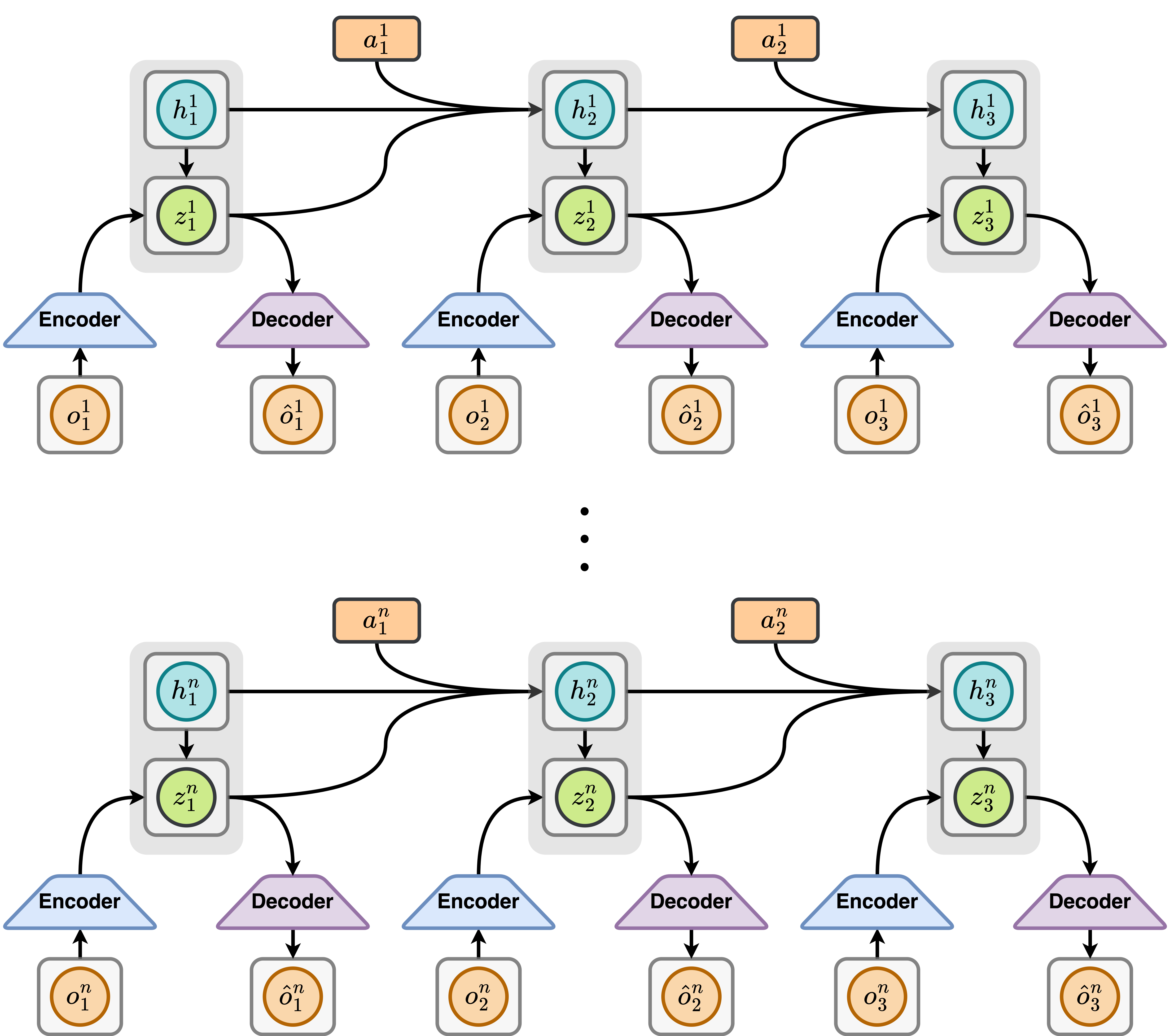}
    \caption[Training process of \acrshort{idreamer} world model]{\textbf{Training process of \acrshort{idreamer} world model:} Each agents' world model encodes their respective observation $o^i_t$ and \acrshort{rssm} recurrent state $h^i_t$ into the discrete stochastic state $z^i_t$ known as the posterior state. Additionally, at each step, the world model for agent $i$ produces a prior state $\hat{z}^i_t$ that is predicted solely by the \acrshort{rssm} recurrent state $h^i_t$ and trained towards $z^i_t$. Given $h^i_t$, $z^i_t$, and the action $a^i_t$, the next recurrent state is produced $h^i_{t+1}$. Each agent's posterior state $z^i_t$ is used to reconstruct the observation $o^i_t$ in order to learn a better representation. The sequence of observations $o^i_{t:T}$ is unrolled over time.}
    \label{fig:idreamer_wm}
\end{figure}

\begin{figure}[h!]
    \centering
    \includegraphics[width=0.65\linewidth]{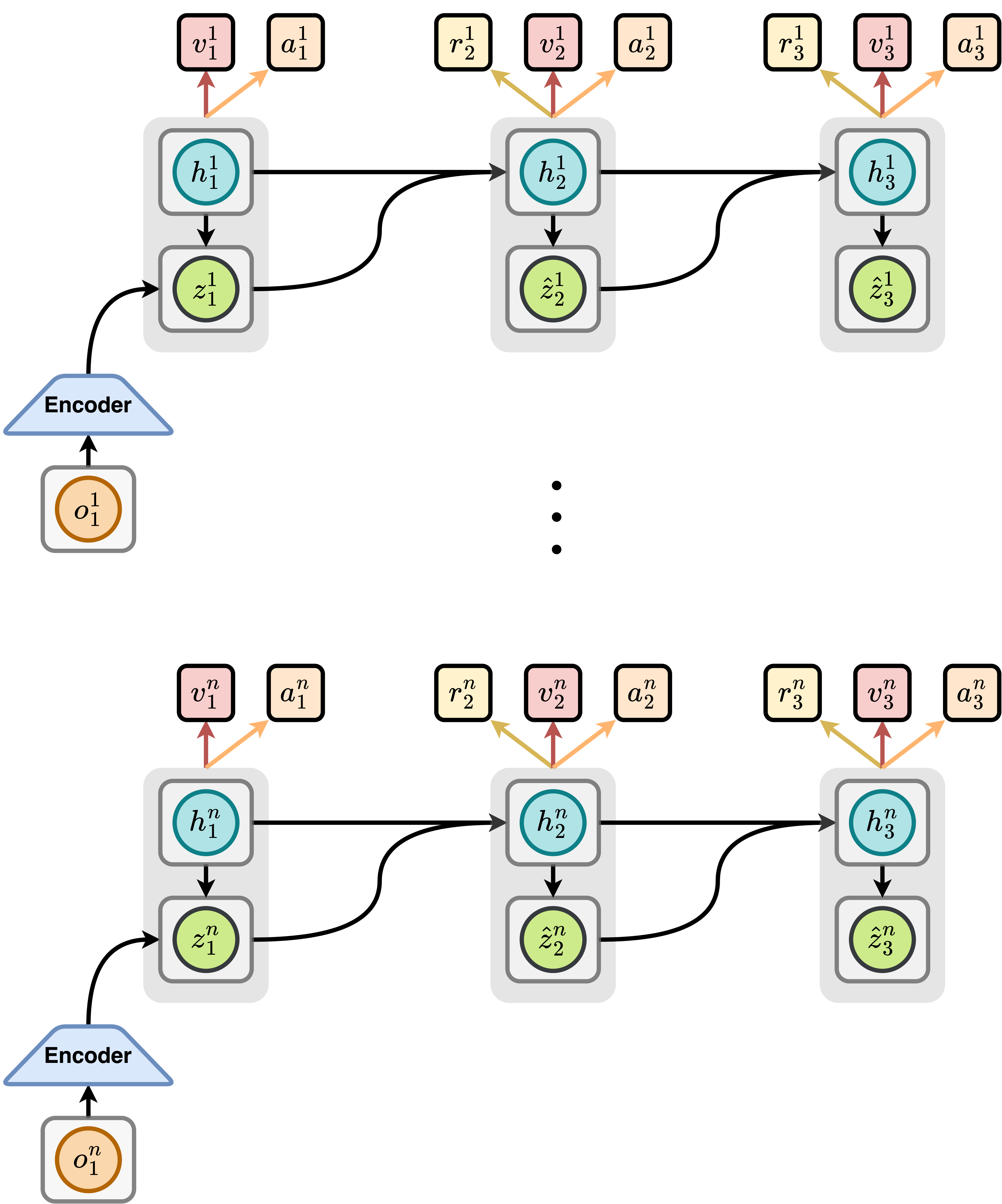}
    \caption[Training process of \acrshort{idreamer} Actor-Critic]{\textbf{Training process of \acrshort{idreamer} Actor-Critic:} Each agents' world model generates synthetic trajectories by auto-regressively predicting the next discrete $\hat{z}^i_t$ and recurrent state $h^i_t$. These states form the compact world model state for each agent's actor-critic networks.}
    \label{fig:idreamer_policy}
\end{figure}

\subsection{Evaluation Methodology}\label{eval-method}

In recent times, there has been a noticeable trend towards evaluating \acrshort{rl} algorithms on larger suites of tasks with comparisons being made on aggregate performance point estimates, such as mean or median, for each task independently. However, these measures often neglect the inherent statistical uncertainty that comes with evaluations based on a limited number of training runs and seeds. Furthermore, a significant portion of research has started to focus on highly computationally-intensive benchmarks whereby each training run can last from a few hours to several weeks. Due to this, it is computationally impractical to evaluate a large number of runs per task thereby further increasing the statistical uncertainty associated with the reported metrics. To mitigate this uncertainty and obtain more statistically validated results, we adopt the evaluation methodology presented in \citet{agarwal2021deep} and \citet{gorsane2022towards} to assess the performance of \acrshort{idreamer} and  \acrshort{codreamer} as well as a model-free baseline I\acrshort{ppo}.

For each algorithm, we conduct evaluations across $M$ tasks within a specific environment suite, utilising $N = 4$ independent training runs per task $m \in M$. During each training run $n \in N$, we assess algorithm performance over $E = 32$ distinct evaluation episodes at intervals of 10,000 environment timesteps. At each interval $i$, we compute the mean return $G^i_{m,n}$ per agent, over the $E$ episodes. Additionally, at each evaluation interval, we employ model checkpointing to save the best-performing model. The model with the largest mean return achieved over all intervals is retained for a subsequent final evaluation.

Upon completing a training run $n \in N$, we evaluate the best model, discovered during interval evaluations, over $10 \times E = 320$ episodes. This evaluation process yields normalised scores $x_{m,n}$, for $m = 1, \ldots, M$ and $n = 1, \ldots, N$, acquired by scaling each per-task score based on the minimum and maximum scores observed throughout all training runs in the specific task. Consequently, we obtain a set of normalised scores $x_{1:M,1:N}$ for each algorithm. To aggregate the performance of an algorithm, we map the set of normalised scores into a singular scalar point estimate i.e. $x_{1:M, 1:N} \rightarrow \Bar{x}$.

By following this methodology, we consolidate all independent tasks $M$ and training runs $N$ of an environment suite into a single comprehensive score and utilise a 95\% \acrfull{ci} obtained from stratified bootstrapping over all $M \times N$ experiments treated as \textit{random samples}. This ensures that although a large number of runs cannot be obtained for each task on its own, by performing statistical analysis over all runs of all tasks, we emulate the statistical confidence obtained when conducting a large number of runs over a single task whilst mitigating a lack of task diversity. Consequently, we report the results for each environment suite as a whole, rather than the tasks independently. 

\subsection{Metrics}
\label{metrics}

Using the set of normalised scores, in addition to traditional aggregate point estimates such as median and mean, we employ the following metrics to measure algorithmic performance:

\begin{itemize}

    \item \textbf{\acrfull{iqm}:} We utilise the \acrshort{iqm}, which ignores both the lower 25\% and upper 25\% of all runs, calculating the mean normalised score of the central 50\% ($\frac{N \cdot M}{2}$). This metric is more robust to outliers in training runs than a conventional \textit{mean} and has less bias when compared to \textit{median}. Additionally, it has been shown that \acrshort{iqm} has greater statistical efficiency than \textit{median} and is thereby able to detect algorithmic improvements using fewer training runs \citep{agarwal2021deep}.
    
    \item \textbf{Probability of Improvement:} The probability of improvement quantifies the likelihood of algorithm $X$ outperforming algorithm $Y$ on a \textbf{randomly} chosen task $m$. This metric is useful in quickly identifying how robust the improvement an algorithm brings. This metric uses the Mann-Whitney U-statistic \citep{mann1947test} for its computation.

    To formally elaborate, we give the following definitions:
    
    \begin{equation}
    Pr(X>Y) = \frac{1}{M}\sum_{m=1}^{M}Pr(X_m > Y_m)
    \end{equation}
    where $Pr(X_m > Y_m)$ is the probability of algorithm $X$ performing better on a \textbf{specific} task $m$.
        
    Furthermore, the metric $Pr(X_m > Y_m)$ is defined as:
    \begin{equation}
    Pr(X_m > Y_m) = \frac{1}{NK}\sum_{i=1}^{N}\sum_{j=1}^{K}S(x_{m,i}, y_{m,j})
    \end{equation}
    where
    \begin{equation}
    S(x, y) = \begin{cases}
        1, & \text{if } y<x \\
        \frac{1}{2}, & \text{if } y=x\\
        0, & \text{if } y>x\\
    \end{cases}
    \end{equation}

    When interpreting the probability of improvement metric, as per the Neyman-Pearson statistical testing criterion outlined by \citet{bouthillier2021accounting}, if the \textbf{lower} bound of the \acrshort{ci} is greater than the null hypothesis of $Pr(X>Y) = 0.5$, then the result is statistically significant. In addition to statistical significance, if the \textbf{upper} bound of the \acrshort{ci} exceeds the recommended threshold of $0.75$, then the result is statistically meaningful. 

    % \item \textbf{Optimality Gap:} The optimality gap is defined as the extent to which an algorithm falls short of achieving a normalised score of $1.0$. As we are using min-max normalised scores, this metric simply measures how far away the aggregated normalised performance is from the maximum score observed during any evaluation episode of a specific task.
\end{itemize}

Additionally, we present algorithm performance using performance profiles, which visually display the entire set of normalised scores $x_{1:M, 1:N}$ and provide enhanced insights into performance variability across tasks compared to interval estimates of aggregate metrics. Specifically, the performance profile reports the fraction of runs that obtained a normalised score above a certain threshold. This allows us to easily perform a visual comparison of methods and determine if one method stochastically dominates another. In the context of performance profiles, if one curve is strictly positioned above another, it is interpreted as having stochastic dominance\footnote{A random variable $X$ is defined as having stochastic dominance over another random variable $Y$ if $P(X > \tau ) \geq P(Y > \tau )$ for all $\tau$, and for at least some $\tau$ there is $P(X >\tau)>P(Y >\tau)$. Intuitively this means that if we sample a random value from $X$, it is likely to be larger than a random value sampled from $Y$.} over the other \citep{levy1992stochastic, dror2019deep, agarwal2021deep}. The performance profiles also allow us to better identify the empirical lower and upper bounds on the performance of an algorithm. Finally, to evaluate sample efficiency, we compute the \acrshort{iqm} score of each interval evaluation and plot it against the number of environment steps taken.

\subsection{Implementation Specifics}

Each algorithm is trained for 500,000 environment timesteps collected by 8 independent workers. All hyperparameters and network sizes are listed in Section \ref{modelsizes} and \ref{hyperparams} in the appendix. Additionally, for both \acrshort{idreamer} and \acrshort{codreamer}, we perform 500 training steps of the world model before any behaviour is learnt. This form of \textit{pre-training} is performed in order to avoid negative performance effects due to a primacy bias, a common \acrshort{rl} flaw examined by \citet{nikishin2022primacy}, when learning on inaccurate world model states. 

As an on-policy algorithm, I\acrshort{ppo} is typically considered to be less sample efficient compared to its off-policy counterparts. However, \acrshort{ppo} makes use of off-policy correction techniques to enable the reuse of recently collected data for multiple training steps. To enhance sample efficiency, we increase the number of epochs and mini-batches within the algorithm, enabling more training steps per batch of collected data. The specific values used can be found in Section \ref{hyperparams}.

\subsection{Model Sizes}
\label{modelsizes}

\begin{table}[h!]
\centering
\begin{tabular}{@{}ll@{}}
\toprule
\textbf{Dimension}                 & \textbf{Size} \\ \midrule
GRU recurrent units       & 512  \\
CNN multiplier            & 32   \\
Dense hidden units        & 512  \\
MLP layers                & 2    \\
RSSM GNN Layers           & 1    \\
Reward \& Cont GNN Layers & 1    \\ \midrule
\textbf{Visual Environment Parameters}       & \textbf{22M}     \\ 
\midrule
\textbf{Vector Environment Parameters}       & \textbf{16M }    \\ 
\bottomrule
\end{tabular}
\caption{World Model Size}
\label{tab:wm-model-size}
\end{table}

\begin{table}[h!]
\centering
\begin{tabular}{@{}ll@{}}
\toprule
\textbf{Dimension}                 & \textbf{Size} \\ \midrule
Dense hidden units        & 512  \\
MLP layers                & 2    \\
GNN Layers                 & 1    \\ 
\midrule
\textbf{Actor Parameters}       & \textbf{1.6M}     \\ \midrule
\textbf{Critic Parameters}       & \textbf{1.7M}    \\ \bottomrule
\end{tabular}
\caption{Actor-Critic Model Size}
\label{tab:actor-critic-model-size}
\end{table}

\begin{table}[h!]
\centering
\begin{tabular}{@{}ll@{}}
\toprule
\textbf{Dimension}                 & \textbf{Size} \\ \midrule
Dense hidden units        & 512  \\
MLP layers                & 2    \\
\midrule
\textbf{Actor Parameters}       & \textbf{1.6M}     \\ \midrule
\textbf{Critic Parameters}       & \textbf{1.7M}    \\ \bottomrule
\end{tabular}
\caption{PPO Model Size}
\label{tab:ppo-model-size}
\end{table}

We list the model sizes used in Tables \ref{tab:wm-model-size}, \ref{tab:actor-critic-model-size}, and \ref{tab:ppo-model-size}.

\subsection{Hyperparameters}
\label{hyperparams}

\begin{table}[h!]
\centering
\begin{tabular}{@{}lll@{}}
\toprule
\textbf{Name}                                & \textbf{Symbol} & \textbf{Value}            \\ \midrule
\textbf{General}                             &        &                  \\ \midrule
Replay capacity (FIFO)              & -      & $10^6$              \\
Batch size                          & B      & 16               \\
Batch length                        & T      & 64               \\
Activation                          & -      & LayerNorm + SiLU \\ \midrule
\textbf{World Model}                         &        &                  \\ \midrule
Number of latents                   & -      & 32               \\
Classes per latent                  & -      & 32               \\
Reconstruction loss scale           & $\beta_{\text{pred}}$  & 1.0              \\
Dynamics loss scale                 & $\beta_{\text{dyn}}$   & 0.5              \\
Representation loss scale           & $\beta_{\text{rep}}$   & 0.1              \\
Learning rate                       & -      & $10^{-4}$             \\
Adam epsilon                        & $\epsilon_a$      & $10^{-8}$             \\
Gradient clipping                   & -      & 1000             \\ \midrule
\textbf{Actor Critic  }                      &        &                  \\ \midrule
Imagination horizon                 & H      & 15               \\
Discount Factor                     & $\gamma$ & 0.997            \\
Return lambda                       & $\lambda$ & 0.95             \\
Target Critic Polyak Averaging Step & -      & 0.02             \\
Actor entropy scale                 & $\eta$      & $3\cdot10^{-4}$           \\
Learning rate                       & -      & $3\cdot10^{-5}$           \\
Adam epsilon                        & $\epsilon_a$      & $10^{-5}$             \\
Gradient clipping                   & -      & 100              \\ \bottomrule
\end{tabular}
\caption{Hyperparameters used in all experiments for both CoDreamer and IDreamer.}
\label{tab:dreamer_hyperparameters}
\end{table}

\begin{table}[h!]
\centering
\begin{tabular}{@{}lll@{}}
\toprule
\textbf{Name}                                & \textbf{Symbol} & \textbf{Value}            \\ \midrule
\textbf{General}                             &        &                  \\ \midrule
Queue capacity (FIFO)              & -       & $1000$              \\
Batch size                          & B      & 64               \\
Batch length                        & T      & 64               \\
Activation                          & -      & ReLu \\ \midrule
\textbf{Algorithm}                         &        &                  \\ \midrule
Learning rate                       & -      & $10^{-3} - 5\cdot10^{-5}$   \\
Learning rate Schedule              & -      & Linear (10,000 training steps)\\
Adam epsilon                        & $\epsilon_a$     & $10^{-5}$             \\
Gradient clipping                   & -      & 1.0             \\ 
Number of Epochs                    & -      & 30              \\
Number of Minibatches               & -      & 32             \\ 
Discount Factor                     & $\gamma$ & 0.997            \\
Clipping Epsilon                    & $\epsilon_c$   &  0.3        \\
Actor entropy scale                 & $\eta$      & $ 10^{-2}$  \\     
Value loss coefficient              & -       & 0.5  \\
GAE lambda                          & $\lambda$ & 0.95             \\

\bottomrule
\end{tabular}
\caption{Hyperparameters used in all experiments for IPPO.}
\label{tab:ippo_hyperparameters}
\end{table}

We list the hyperparameters used in Tables \ref{tab:dreamer_hyperparameters} and \ref{tab:ippo_hyperparameters}.

\subsection{Evaluation Details}

As recommended by \citet{agarwal2021deep}, for our aggregate point estimates confidence intervals, we use stratified bootstrapping with 50,000 bootstrap replications. For our probability of improvement and performance profiles, we use 2000 bootstrap replications. Lastly, for our sample efficiency plots, we use 5000 bootstrap replications. 

\begin{table}[h!]
\resizebox{\textwidth}{!}{
\begin{tabular}{@{}llllcc@{}}
\toprule
\textbf{Environment} & \textbf{Task}               & \textbf{Minimum Observed Return}  & \textbf{Maximum Observed Return} \\ \midrule
Estimate Game & Sequential & -3.13 &  0.00 \\
\acrshort{vmas}        & Flocking           &   -37.65      &  2.58 \\
\acrshort{vmas}        & Discovery          &    0.00      &  14.20\\
\acrshort{vmas}        & Buzz Wire          &    -29.97     &  1.81\\
Melting Pot  & Daycare              &  0 &  189.00 \\
Melting Pot  & Cooperative Mining & 8.00 &  1003.00 \\
Melting Pot  & Collaborative Cooking: Asymmetric & 0.00 & 2514.00 \\            
Melting Pot  & Collaborative Cooking: Forced & 0.00 & 49.00 \\\bottomrule
\end{tabular}
}
\caption{Observed Min-Max scores used for normalisation}
\label{tab:min-max-scores}
\end{table}

We list the minimum and maximum scores observed and used for normalisation in Table \ref{tab:min-max-scores}.

\subsection{Evaluation Environments Description}

\begin{table}[h!]
\resizebox{\linewidth}{!}{
\begin{tabular}{@{}llllcc@{}}
\toprule
\textbf{Environment} & \textbf{Task}               & \textbf{Observation Spec.}  & \textbf{Action Spec.} & \textbf{No. of Agents} & \textbf{Avg. Time Horizon} \\ \midrule
\acrshort{vmas}        & Flocking           & Vector (18)         & Discrete (5) & 4                & 500                    \\
\acrshort{vmas}        & Discovery          & Vector (21)         & Discrete (5) & 5                & 500                    \\
\acrshort{vmas}        & Buzz Wire          & Vector (8)         & Discrete (5) & 2                & 500                    \\
Melting Pot  & Daycare              & Pixels (64, 64, 3) & Discrete (8) & 2                & 1000                   \\
Melting Pot  & Cooperative Mining & Pixels (64, 64, 3) & Discrete (8) & 4                & 1500                   \\
Melting Pot  & Collaborative Cooking: Asymmetric & Pixels (64, 64, 3) & Discrete (8) & 2                & 1000       \\            
Melting Pot  & Collaborative Cooking: Forced & Pixels (64, 64, 3) & Discrete (8) & 2                & 1000\\\bottomrule
\end{tabular}
}
\caption{External Environment Summary}
\label{tab:ext_env_summary}
\end{table}

\subsubsection{Flocking}

\begin{figure}[h!]
    \centering
    \includegraphics[width=\linewidth]{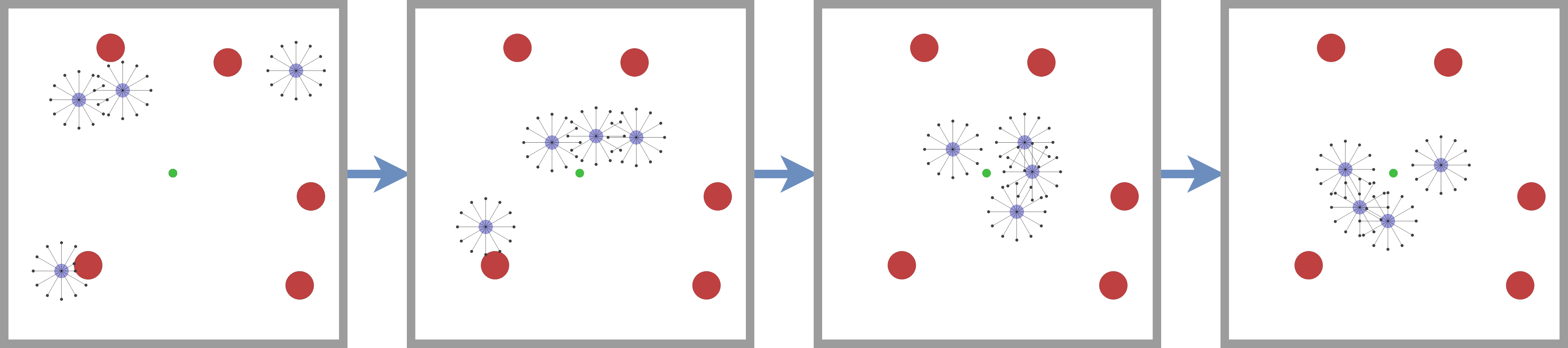}
    \caption[\acrshort{vmas} Flocking environment]{\textbf{\acrshort{vmas} Flocking} environment. Example illustration showcasing 4 agents surrounding a target position whilst avoiding obstacles}
    \label{fig:flocking-illustration}
\end{figure}

The Flocking task (see Figure \ref{fig:flocking-illustration}) generates an open environment containing $M$ randomly positioned obstacles. $N$ agents are situated within this environment, tasked with encircling a moving target whilst avoiding collisions with the obstacles as well as one another. Flocking has been a long-standing benchmark in the field of robotic coordination \citep{reynolds1987flocks} and serves as an ideal challenge due to the complexity of coordinating multiple agents. We specify our instantiation of the Flocking task in Table \ref{tab:flocking-specifics}.

\begin{table}[h!]
\centering
\begin{tabular}{@{}lcc@{}}
\toprule
\textbf{Name}                   & \textbf{Symbol}           & \textbf{Value}   \\ \midrule
Number of Agents       & $N$              & 4       \\
Number of Obstacles   & $M$              & 5       \\
\bottomrule
\end{tabular}
\caption{Specific instantiation of \acrshort{vmas} Flocking}
\label{tab:flocking-specifics}
\end{table}

\subsubsection{Discovery}

\begin{figure}[h!]
    \centering
    \includegraphics[width=\linewidth]{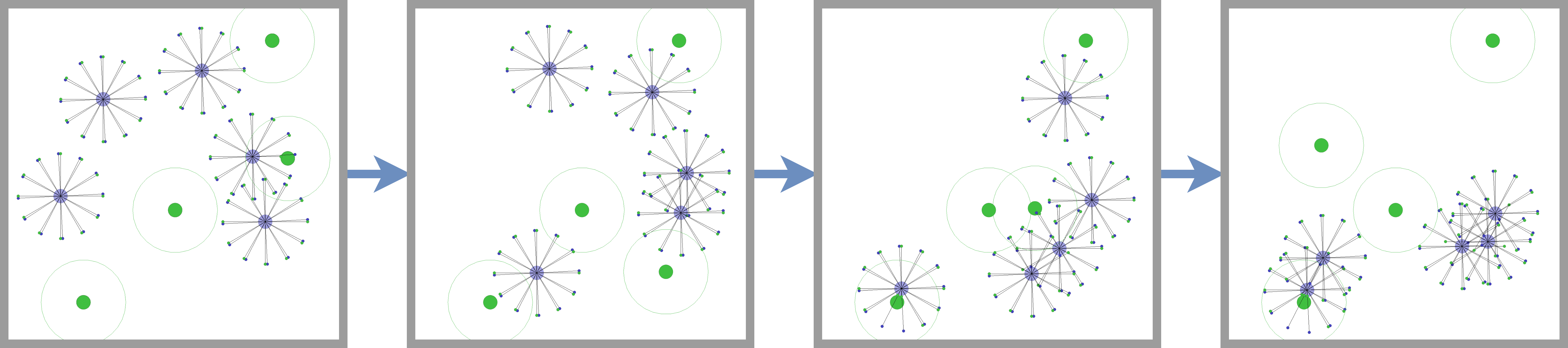}
    \caption[\acrshort{vmas} Discovery environment]{\textbf{\acrshort{vmas} Discovery} environment. Example illustration showcasing 5 agents covering 7 objectives.}
    \label{fig:discovery-illustration}
\end{figure}

The Discovery task, drawing inspiration from the Stick Pulling experiment \citep{ijspeert2001collaboration}, places $N$ agents in an open setting with $M$ objectives. The agents are assigned to cover as many objectives as they can while avoiding collisions. An objective is considered satisfied when $K$ agents are within a predetermined distance $D$ from it. After a goal has been covered by $K$ agents, each of them is rewarded. It has been shown that the performance can be significantly improved through communication when the number of agents, $N$, is less than the number of goals, $M$. We specify our instantiation of the Discovery task in Table \ref{tab:discovery-specifics}.

\begin{table}[h!]
\centering
\begin{tabular}{@{}lcc@{}}
\toprule
\textbf{Name}                   & \textbf{Symbol}           & \textbf{Value}   \\ \midrule
Number of Agents       & $N$              & 5       \\
Number of Objectives   & $M$              & 7       \\
Coverage Requirement   & $K$              & 2       \\
Coverage Distance      & $D$              & 0.25    \\
\bottomrule
\end{tabular}
\caption{Specific instantiation of \acrshort{vmas} Discovery}
\label{tab:discovery-specifics}
\end{table}

\subsubsection{Buzz Wire}

\begin{figure}[h!]
    \centering
    \includegraphics[width=\linewidth]{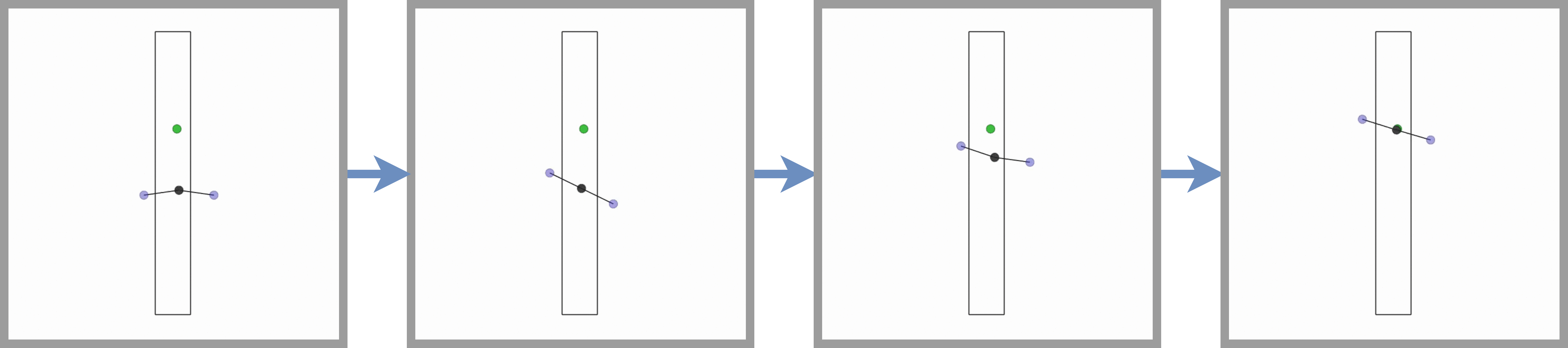}
    \caption[\acrshort{vmas} Buzz Wire environment]{\textbf{\acrshort{vmas} Buzz Wire} environment. Example illustration showcasing 2 agents successfully reaching the target.}
    \label{fig:buzzwire-illustration}
\end{figure}

The Buzz Wire task, based on the popular ``Wire Loop" game \citep{buzzwire_article}, requires two agents to steer a central mass via attached linkages through a straight corridor to a designated target. Both agents are unable to touch the borders of the corridor otherwise they fail the task and the episode ends, introducing a fair level of difficulty. This task necessitates a high level of awareness and coordination between the agents, as a lack of synchronised movement can result in one agent pulling the other into the borders thereby causing failure.

\subsubsection{Daycare}

\begin{figure}[h!]
    \centering
    \includegraphics[width=\linewidth]{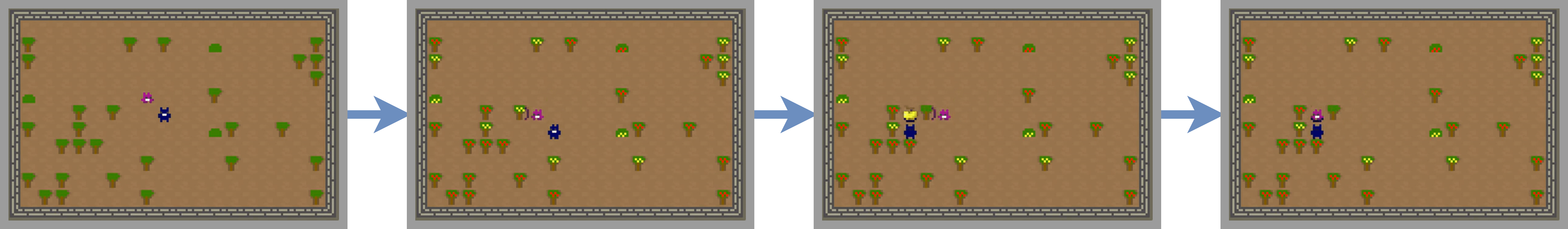}
    \caption[Melting Pot: Daycare environment]{\textbf{Melting Pot: Daycare} environment. Example illustration the parent and child agents collecting food.}
    \label{fig:Daycare-illustration}
\end{figure}

Daycare is a straightforward two-player game in which agents occupy one of two roles: \textit{parent} or \textit{child}. In Daycare, two types of fruit can grow on either shrubs or trees. The parent can pick and consume any fruit from a tree or shrub whereas the child can only pick fruits on shrubs and only consumes one specific type of fruit. Each agent is rewarded equally for consuming fruit. If the child does not consume a fruit within 200 timesteps, it throws a ``tantrum" and is temporarily removed from the game for 100 timesteps. Whilst the child is removed, the parent cannot gain any reward, thus, two challenges exist: First, the parent must take advantage of its unique affordances and assist the child to pick the specific type of fruit for it from trees. Second, the child needs to convey its fruit preference to the parent as it can change from episode to episode.

\subsubsection{Cooperative Mining}

\begin{figure}[h!]
    \centering
    \includegraphics[width=\linewidth]{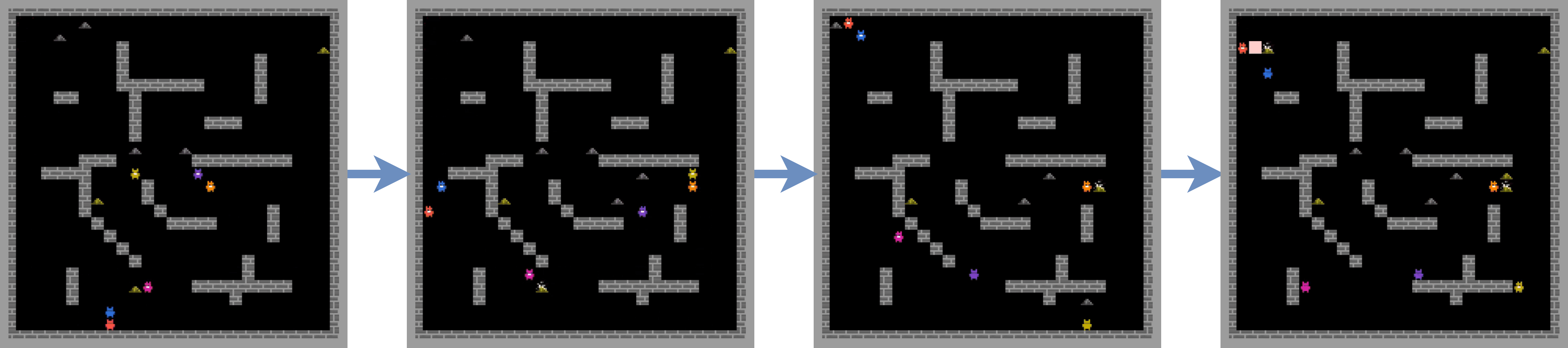}
    \caption[Melting Pot: Cooperative Mining environment]{\textbf{Melting Pot: Cooperative Mining} environment. Example illustration showcasing four agents working together to mine Iron and Gold ores.}
    \label{fig:coopmining-illustration}
\end{figure}

Cooperative Mining is originally a six-player game designed to assess the effectiveness of cooperation in multi-agent systems. In this game, two types of ore, Iron and Gold, spawn randomly in empty spaces throughout the map. When an agent mines Iron ore, they receive a reward of 1, and this extraction process requires no collaboration. In contrast, Gold ore demands the coordinated efforts of two agents for successful mining, granting a reward of 8 to each participant. When an agent initiates mining gold, the ore flashes, signalling other agents to mine within a 3-timestep window. If no other agent attempts mining or too many agents engage, the gold ore reverts to its regular state and no rewards are given.

By encouraging agents to maintain close proximity and collaborate in mining gold, higher rewards can be achieved compared to mining iron individually. Additionally, through the communication of ore locations and mining intentions, the global reward of the system can be further increased. Thus, the Cooperative Mining game serves as a valuable benchmark for evaluating cooperation and coordination in multi-agent learning algorithms. In this work, we utilise only four agents instead of six.

\subsubsection{Collaborative Cooking}

Based on the game \textit{Overcooked} \citep{overcooked}, Melting Pot's Collaborative Cooking is a unique environment where multiple players work together to follow recipes and serve food to customers. This task involves agents creating soup by cooking three tomatoes in a pot for 20 timesteps, followed by plating and serving to customers. Depending on the map configuration and the number of players, these scenarios can demand a high degree of collaboration and cooperation to achieve sufficient performance. In this work, we focus on two specific scenarios: \textbf{Asymmetric} and \textbf{Forced}.

\subsubsection{Asymmetric}

\begin{figure}[h!]
    \centering
    \includegraphics[width=\linewidth]{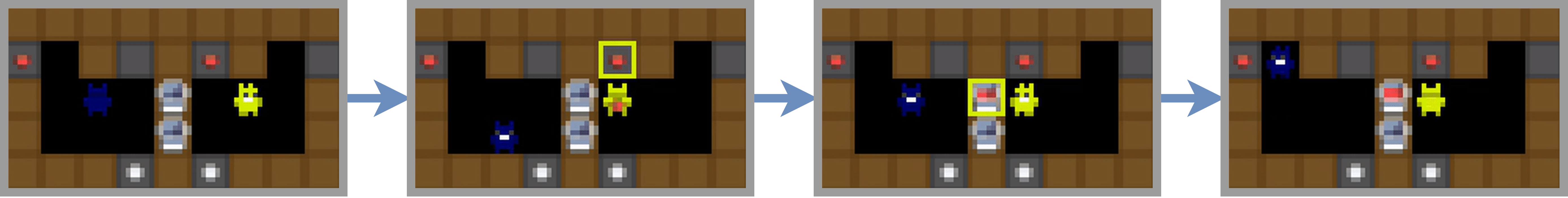}
    \caption[Melting Pot: Collaborative Cooking - Asymmetric environment]{\textbf{Melting Pot: Collaborative Cooking - Asymmetric} environment. Example illustration of the asymmetric scenario's map layout.}
    \label{fig:colabcooking-asymmetric-illustration}
\end{figure}

In the Asymmetric scenario, two players are separately placed in non-connected parts of the map with differing proximity to plates, pots, and tomatoes. For one player, the goal delivery spot is located close to the cooking pots, but far from the tomato dispenser. Conversely, the second player has the goal delivery spot far from their pots, but close to the tomato dispenser. This setup allows both players to function independently to some extent, but they can achieve a significantly greater reward if they learn to collaborate and leverage their individual advantages by specialising to their side of the map.

\subsubsection{Forced}

\begin{figure}[h!]
    \centering
    \includegraphics[width=\linewidth]{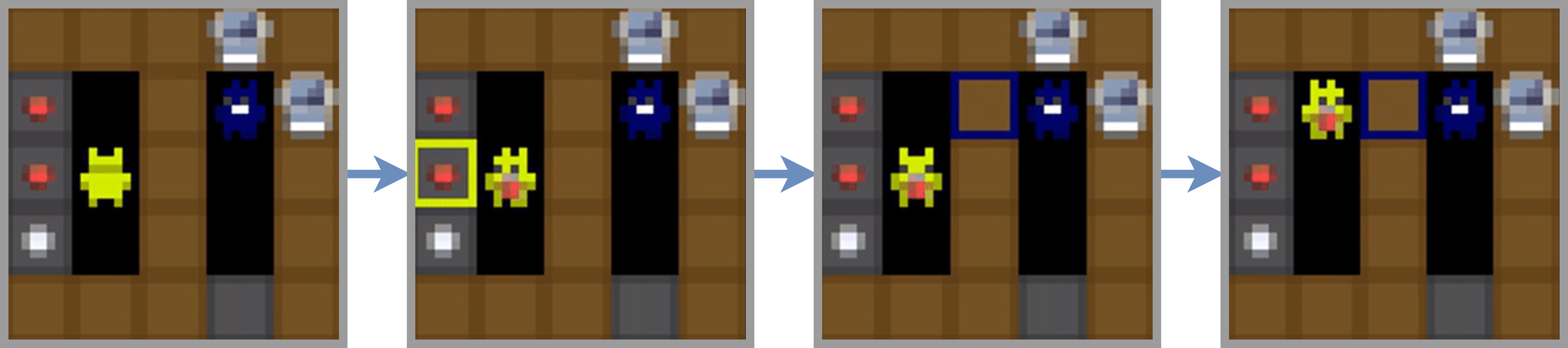}
    \caption[Melting Pot: Collaborative Cooking - Forced environment]{\textbf{Melting Pot: Collaborative Cooking - Forced} environment. Example illustration of the Forced scenario's map layout.}
    \label{fig:colabcooking-forced-illustration}
\end{figure}

The Forced scenario, like the Asymmetric one, features two players on separate and non-connected parts of the map. However, differing from Asymmetric, each player has exclusive access to certain resources. One player has access to the tomato and plate dispensers, while the other has access to the cooking pots and the drop-off point. Thus, to earn any reward in this setup, the players must work together and cooperate effectively.

\subsection{Environmental Impact}

In the final evaluation of this work, we ran 5 different algorithms: I\acrshort{ppo}, \acrshort{idreamer}, \acrshort{codreamer}, AC Comm, and WM Comm. Each of these algorithms was evaluated for 4 runs on 7 tasks each. This equates to $5 \times 4 \times 7 = 140$ experimental training runs. Each training run lasted for approximately $6$ to $12$ hours. Taking the average experiment time of $9$ hours, we utilised hardware accelerators for a total of $140 \times 9 = 1260$ hours.

All experiments were conducted using TPU v2 and v3 chips in the Europe-west4 and us-central1 regions on the Google Cloud Platform, which has an approximate carbon efficiency of 0.57 kgCO$_2$eq/kWh. The TPUv2 and v3 chips have an approximate power draw of 221W and 283W respectively. Using the average between these, we set the power draw of the hardware used to 252W. 

Given these values, total carbon emissions are estimated to be:
\[252\text{W} \times 1260\text{h} = 317.52\text{kWh} \times 0.57\text{ kgCO$_2$eq/kWh} = 180.99 \text{ kgCO$_2$eq}\]

These estimates are exclusively calculated for the final evaluation and not for any experiments run during the development of the proposed methods. 
    
These estimations were in part conducted using the \href{https://mlco2.github.io/impact#compute}{MachineLearning Impact calculator} as presented in \citet{lacoste2019quantifying}.

\subsection{Supplementary Evaluation Information}

\begin{table}[h!]
    \centering
\begin{tabular}{@{}lccc@{}}
\toprule
 & \textbf{IDREAMER} & \textbf{IPPO} & \textbf{CODREAMER} \\
 \midrule
\textbf{Median} & 0.89 [0.85, 0.91] & 0.80 [0.78, 0.82] & 0.97 [0.96, 0.97] \\
\textbf{IQM} & 0.90 [0.84, 0.91] & 0.79 [0.78, 0.83] & 0.97 [0.96, 0.97] \\
\textbf{Mean} & 0.89 [0.85, 0.91] & 0.80 [0.78, 0.82] & 0.97 [0.96, 0.97] \\
\textbf{Optimality Gap} & 0.11 [0.09, 0.15] & 0.20 [0.18, 0.22] & 0.03 [0.03, 0.04] \\
\bottomrule
\end{tabular}
    \caption[Aggregated Results for Sequential Estimate Game]{Aggregated Results for \textbf{Sequential Estimate Game}}
    \label{tab:agg_estimate_results}
\end{table}

\begin{table}[h!]
    \centering
    \begin{tabular}{@{}lccc@{}}
    \toprule
 & \textbf{IDREAMER} & \textbf{IPPO} & \textbf{CODREAMER} \\
 \midrule
\textbf{Median} & 0.91 [0.90, 0.92] & 0.92 [0.92, 0.93] & 0.93 [0.93, 0.94] \\
\textbf{IQM} & 0.81 [0.79, 0.82] & 0.79 [0.78, 0.79] & 0.84 [0.82, 0.84] \\
\textbf{Mean} & 0.68 [0.66, 0.71] & 0.65 [0.65, 0.66] & 0.72 [0.70, 0.74] \\
\textbf{Optimality Gap} & 0.32 [0.29, 0.34] & 0.35 [0.34, 0.35] & 0.28 [0.26, 0.30] \\
\bottomrule
\end{tabular}
    \caption[Aggregated Results for \acrshort{vmas}]{Aggregated Results for \textbf{VMAS}}
    \label{tab:agg_vmas_results}
\end{table}

\begin{table}[h!]
    \centering
    \begin{tabular}{@{}lccc@{}}
    \toprule
 & \textbf{IDREAMER} & \textbf{IPPO} & \textbf{CODREAMER} \\
 \midrule
\textbf{Median} & 0.30 [0.17, 0.44] & 0.05 [0.05, 0.06] & 0.48 [0.45, 0.50] \\
\textbf{IQM} & 0.30 [0.17, 0.44] & 0.05 [0.05, 0.06] & 0.48 [0.45, 0.50] \\
\textbf{Mean} & 0.34 [0.25, 0.44] & 0.05 [0.04, 0.05] & 0.51 [0.50, 0.53] \\
\textbf{Optimality} Gap & 0.66 [0.56, 0.75] & 0.95 [0.95, 0.96] & 0.49 [0.47, 0.50] \\
\bottomrule
\end{tabular}
    \caption[Aggregated Results for Melting Pot]{Aggregated Results for \textbf{Melting Pot}}
    \label{tab:agg_meltingpot_results}
\end{table}

\begin{figure}
    \centering
    \includegraphics[width=\linewidth]{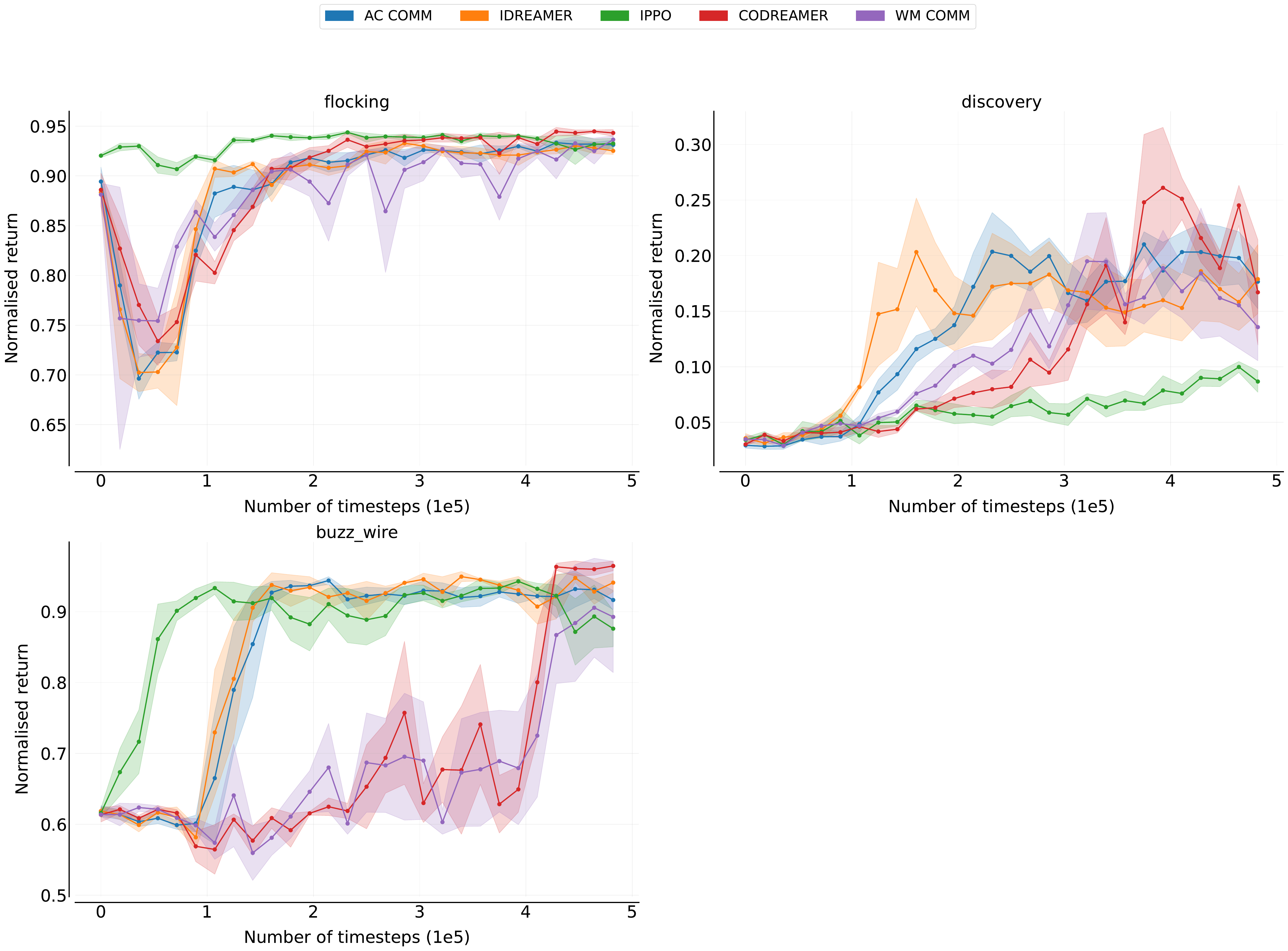}
    \caption[Individual IQM plots for each task in \acrshort{vmas}]{Individual IQM plots for each task in \textbf{VMAS}}
    \label{fig:vmas-single-tasks}
\end{figure}

\begin{figure}
    \centering
    \includegraphics[width=\linewidth]{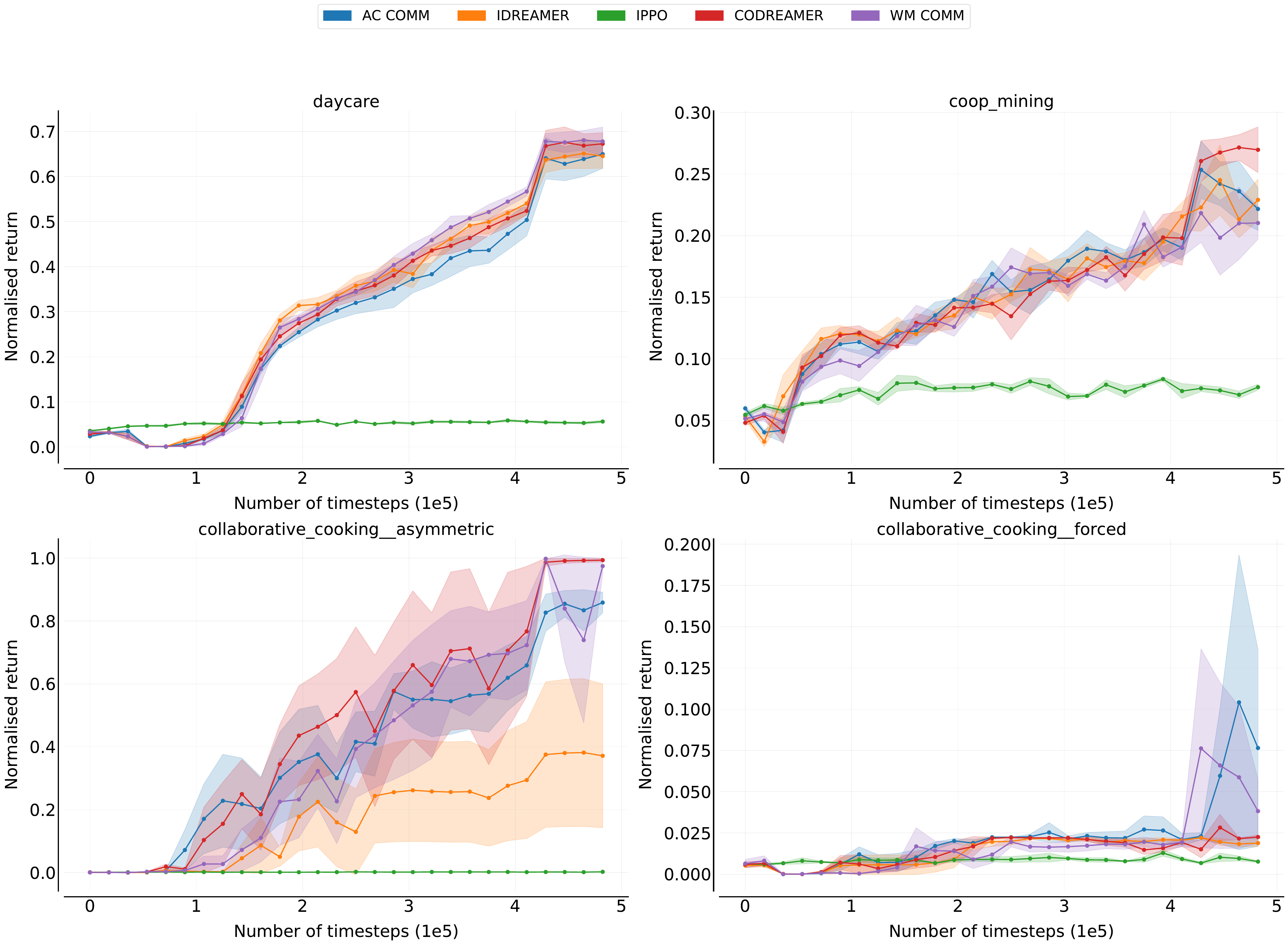}
    \caption[Individual IQM plots for each task in Melting Pot]{Individual IQM plots for each task in \textbf{Melting Pot}}
    \label{fig:meltingpot-single-tasks}
\end{figure}

We list the results obtained in Table \ref{tab:agg_estimate_results}, \ref{tab:agg_vmas_results}, and \ref{tab:agg_meltingpot_results}. Additionally, we present the plots for each task in Figures \ref{fig:vmas-single-tasks} and \ref{fig:meltingpot-single-tasks}.

\end{document}